\documentclass{article}

\usepackage{microtype}
\usepackage{graphicx}
\usepackage{booktabs} 

\usepackage{hyperref}

\usepackage[accepted]{icml2023}

\usepackage{amsmath}
\usepackage{amssymb}
\usepackage{mathtools}
\usepackage{amsthm}

\usepackage[capitalize,noabbrev]{cleveref}

\theoremstyle{plain}

\theoremstyle{definition}

\theoremstyle{remark}

\usepackage[textsize=tiny]{todonotes}

\usepackage{colortbl}
\usepackage{xcolor}
\usepackage{enumitem}
\usepackage{multirow}
\usepackage{pifont}
\usepackage{caption}
\usepackage{subcaption}
\usepackage{lipsum}

\icmltitlerunning{For Pre-Trained Vision Models in Motor Control, Not All Policy Learning Methods are Created Equal}

\begin{document}

\twocolumn[
\icmltitle{For Pre-Trained Vision Models in Motor Control, \\ Not All Policy Learning Methods are Created Equal}



\icmlsetsymbol{equal}{*}
\icmlsetsymbol{advising}{$\dag$}

\begin{icmlauthorlist}
\icmlauthor{Yingdong Hu}{thu,shlab,qz}
\icmlauthor{Renhao Wang}{thu}
\icmlauthor{Li Erran Li}{aws,advising}
\icmlauthor{Yang Gao}{thu,shlab,qz}
\end{icmlauthorlist}

\icmlaffiliation{thu}{Tsinghua University}
\icmlaffiliation{shlab}{Shanghai Artificial Intelligence Laboratory}
\icmlaffiliation{qz}{Shanghai Qi Zhi Institute}
\icmlaffiliation{aws}{AWS AI, Amazon}

\icmlcorrespondingauthor{Yang Gao}{gaoyangiiis@tsinghua.edu.cn}
\icmlcorrespondingauthor{Yingdong Hu}{huyd21@mails.tsinghua.edu.cn}




\icmlkeywords{Pre-trained Vision Models, Motor Control}

\vskip 0.3in
]



\printAffiliationsAndNotice{}  

\begin{abstract}
In recent years, increasing attention has been directed to leveraging pre-trained vision models for motor control. While existing works mainly emphasize the importance of this pre-training phase, the arguably equally important role played by downstream policy learning during control-specific fine-tuning is often neglected. It thus remains unclear if pre-trained vision models are consistent in their effectiveness under different control policies. To bridge this gap in understanding, we conduct a comprehensive study on 14 pre-trained vision models using 3 distinct classes of policy learning methods, including reinforcement learning (RL), imitation learning through behavior cloning (BC), and imitation learning with a visual reward function (VRF). Our study yields a series of intriguing results, including the discovery that the effectiveness of pre-training is highly dependent on the choice of the downstream policy learning algorithm. We show that conventionally accepted evaluation based on RL methods is highly variable and therefore unreliable, and further advocate for using more robust methods like VRF and BC. To facilitate more universal evaluations of pre-trained models and their policy learning methods in the future, we also release a benchmark of 21 tasks across 3 different environments alongside our work.
Source code and more details can be found at \url{https://yingdong-hu.github.io/PVM-control/}.
\end{abstract}

\section{Introduction}
The transfer of pre-trained features to various parallel and even orthogonal downstream tasks is a ubiquitous paradigm in modern deep learning. This well-explored approach has revolutionized computer vision~\cite{chen2020simple,he2020momentum}, natural language processing~\cite{devlin2018bert,brown2020language}, and other fields~\cite{baevski2020wav2vec}. The transferability of visual features has been thoroughly demonstrated~\cite{tan2018survey}, leading to an explosion of pre-trained models in computer vision.

\begin{figure}[t!]
    \centering
    \includegraphics[width=1.0\columnwidth]{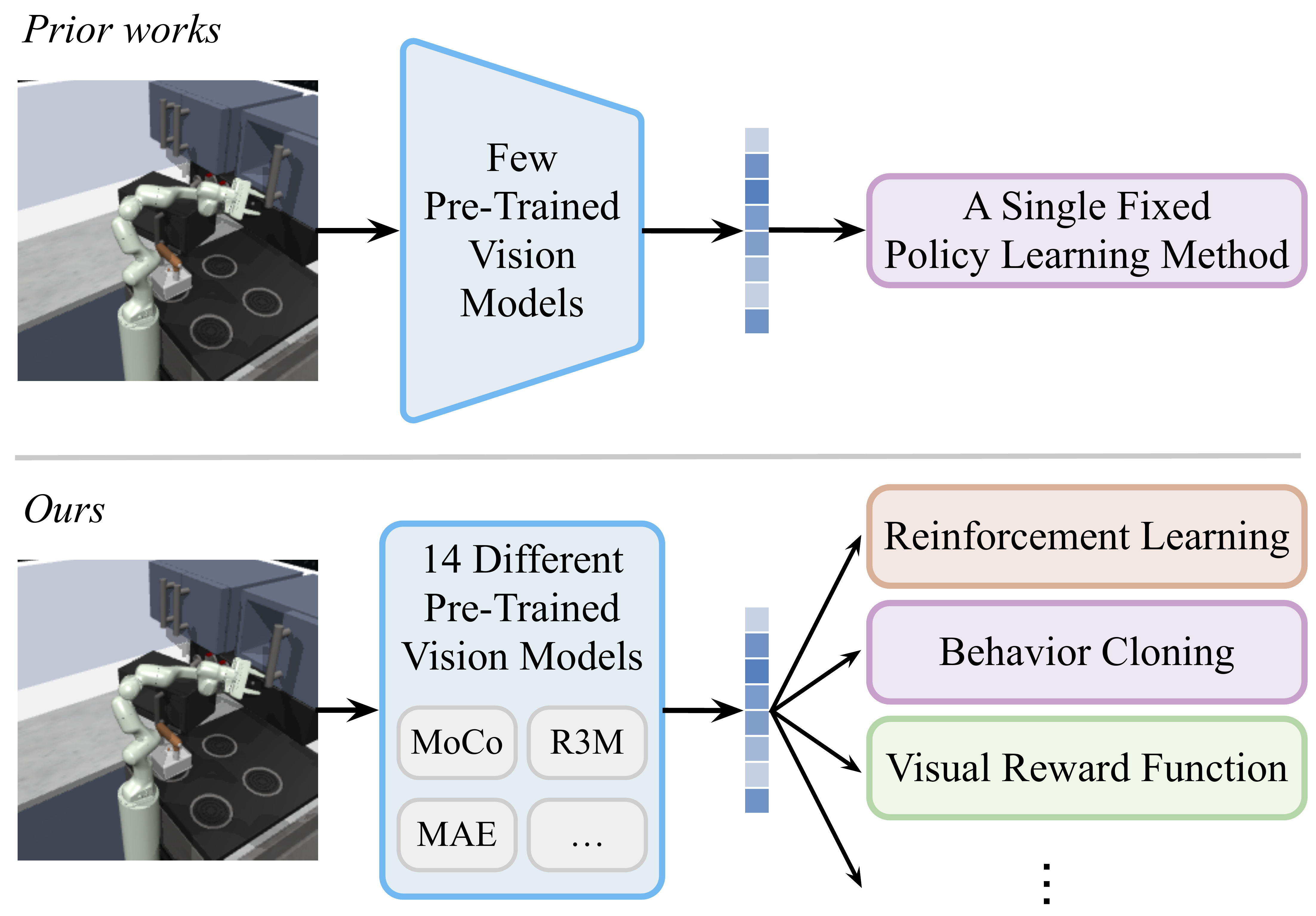} 
    \caption{\textbf{(Top)} Most prior works that leverage pre-trained vision models as frozen perception modules for motor control only compare a few models using a single fixed policy learning algorithm. \textbf{(Bottom)} We find that using different policy learning algorithms results in significant changes in the rankings of 14 different vision models, i.e., the effectiveness of a vision model is algorithm-dependent.}
    \label{fig:teaser}
    \vspace{-1.5em}
\end{figure}

A growing body of research is focused on rendering such pre-trained models as the cornerstones for vision-based motor control~\cite{parisi2022unsurprising,xiao2022masked,radosavovic2022real,nair2022r3m,ma2022vip}. Since these models have typically been trained on large-scale natural visual data, their features possess general knowledge about the semantics of our world and its properties, which is invaluable for universal control. A downstream set of modules then learn a control policy by adapting these out-of-domain features to in-domain control data.

However, the emphasis to date has largely been on how the pre-training phase can be done better. To learn a downstream control policy, prior works often make ad-hoc choices around reinforcement learning or imitation learning approaches~\cite{xiao2022masked, parisi2022unsurprising}, resulting in a lack of understanding of the impact of downstream control policy on final performance. Can a given pre-trained vision model maintain consistent effectiveness across different downstream policy learning methods? If not, can we explain the difference? Given multiple possible downstream policy learning methods, how should we evaluate a pre-trained vision model?

To answer these questions, we conduct a large-scale benchmarking study on diverse pre-trained vision models for a plethora of control tasks across different environments. Our study treats the often-neglected policy learning phase as a first-class citizen. As shown in Figure~\ref{fig:teaser}, we consider \textit{three} policy learning algorithms: (\textit{i}) reinforcement learning (RL), (\textit{ii}) imitation learning through behavior cloning (BC), and (\textit{iii}) imitation learning with a visual reward function (VRF). 
The first two approaches (RL and BC) are widely used in the existing literature and treat pre-trained features as representations that encode environment-related information. The last approach (VRF) is an inverse reinforcement learning (IRL) paradigm we adopt which requires that the pre-trained features also capture a high-level notion of task progress, an idea that remains largely underexplored. We then consider 14 pre-trained vision models covering different architecture
(ResNet;~\citealt{he2016deep} and ViT;~\citealt{dosovitskiy2020image}) and prevalent pre-training methods (contrastive learning;~\citealt{chen2021empirical}, self-distillation;~\citealt{caron2021emerging}, language-supervised;~\citealt{radford2021learning}, masked image modeling;~\citealt{bao2021beit}, etc.). For a fair and comprehensive comparison, we run extensive experiments on 21 simulated tasks across 3 robot manipulation environments: Meta-World~\cite{yu2020meta}, Franka-Kitchen~\cite{gupta2019relay}, and Robosuite~\cite{zhu2020robosuite}. Our investigation reveals surprising results and contributions:

\begin{itemize}[topsep=0pt, partopsep=0pt, leftmargin=13pt, parsep=0pt, itemsep=4pt]
    \item \textbf{Lack of consistently performant models.} The effectiveness of a pre-trained vision model is highly dependent on the downstream policy learning method.

    \item \textbf{Point out directions for reliable evaluation methods.} 
    Due to high variability, RL is not a robust evaluation method. We show that the consistent results of VRF and BC make them reliable evaluation methods in our benchmark of pre-trained vision models for motor control.

    \item \textbf{Deeper dive into properties of vision models} enables us to obtain metrics, such as linear probing loss and $k$-NN classification accuracy, that have substantive predictive power for downstream control policies.
    
\end{itemize}
\section{Related Work}

\textbf{Pre-training in computer vision.}
Large-scale pre-training has become the new fuel empowering computer vision.
Contrastive learning and related methods~\cite{hadsell2006dimensionality,wu2018unsupervised,caron2020unsupervised,hu2022semantic} learn visual representations by modeling image similarity~\cite{grill2020bootstrap} and dissimilarity~\cite{chen2020simple} between two or more views. 
Masked Image Modeling (MIM)~\cite{bao2021beit} pursues a different direction by learning to predict removed pixels~\cite{he2022masked}, discrete visual tokens~\cite{peng2022beit}, or pre-computed features~\cite{wei2022masked}.
Language-supervised pre-training, e.g., CLIP~\cite{radford2021learning} and related works~\cite{mu2022slip,dong2022maskclip}, has been established as a powerful paradigm for learning visual representations. 
While pre-trained models attract increasing attention in the vision field, no large-scale evaluation has compared the various models available for motor control. This work aims to benchmark the plethora of pre-trained vision models to explore which ones are the most effective for visuomotor control.

\textbf{Pre-trained vision models for motor control.}
The application of pre-trained vision models to problems in motor control is a rapidly growing field~\cite{radosavovic2022real,wang2022vrl3}, with studies such as RRL~\cite{shah2021rrl}, PIE-G~\cite{yuan2022pre}, and MVP~\cite{xiao2022masked} demonstrating the effectiveness of supervised or self-supervised pre-trained vision models as visual representations for RL agents.
PVR~\cite{parisi2022unsurprising} and R3M~\cite{nair2022r3m} find that vision models pre-trained on real-world data enable data-efficient behavior cloning on diverse control tasks.
VIP~\cite{ma2022vip} proposes a self-supervised pre-trained vision model capable of producing dense reward signals. Concurrently, Hansen et al.~\yrcite{hansen2022pre} show that a carefully designed Learning-from-Scratch (LfS) baseline is competitive with methods that leverage pre-trained vision models. 
However, most approaches train the agent with only BC or only RL, with limited or no discussion on how policy learning choices are made. Thus, it remains unclear whether the effectiveness of pre-trained vision models is consistent across different policy learning methods.

\begin{figure*}[t!]
    \centering
    \includegraphics[width=1.0\linewidth]{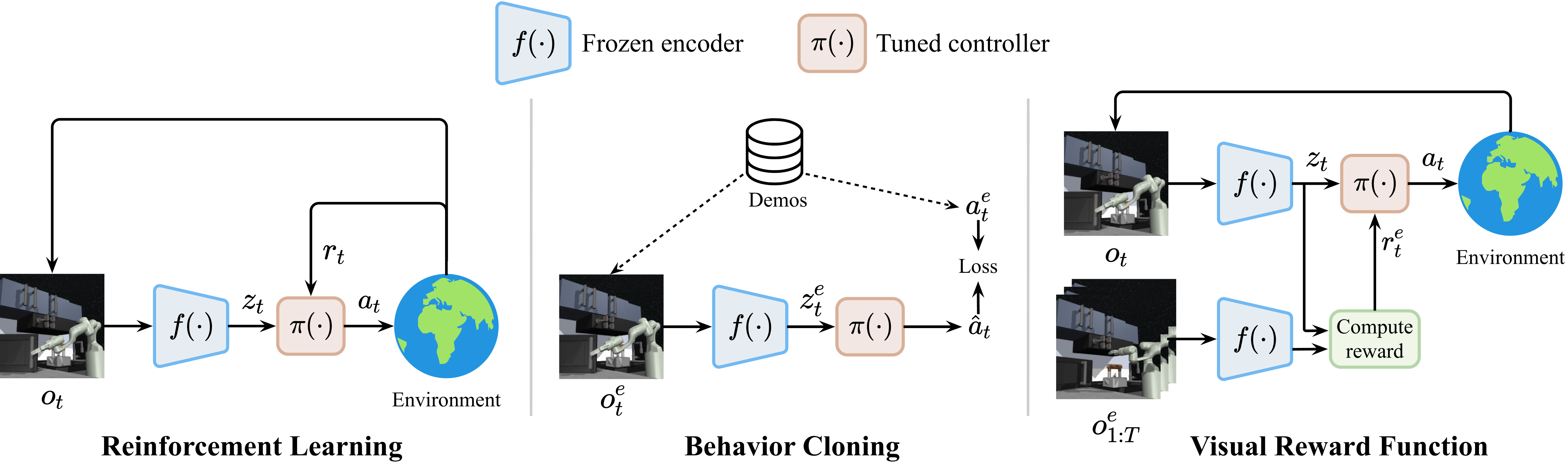} 
    \caption{Illustration of three downstream policy learning methods we considered. From left to right: reinforcement learning (RL), imitation learning through behavior cloning (BC), and imitation learning with a visual reward function (VRF).}
    \label{fig:policy_learning_methods}
    \vspace{-1em}
\end{figure*}

\textbf{Policy learning.}
Reinforcement learning (RL)~\cite{sutton2018reinforcement} and imitation learning (IL)~\cite{hussein2017imitation} are two mainstream approaches for policy learning. 
The gap between image-based RL and state-based RL has been significantly bridged, largely due to ideas like autoencoder-based architectures~\cite{hafner2019learning,yarats2021improving}, self-supervised objectives~\cite{laskin2020curl,schwarzer2020data}, and data augmentation~\cite{kostrikov2020image,laskin2020reinforcement}.
IL can be broadly categorized into Behavior Cloning (BC)~\cite{pomerleau1988alvinn} and Inverse Reinforcement Learning (IRL)~\cite{ng2000algorithms}. 
BC is extremely sample-efficient but may suffer on out-of-distributions samples ~\cite{ross2011reduction} or copycat problems~\cite{wen2020fighting}. IRL focuses on learning a robust reward function~\cite{kostrikov2018discriminator}.
In this work, we aim to contrast the merits of three different policy learning methods (i.e., RL, BC, IRL) and their properties with respect to re-appropriating pre-trained general vision models for downstream control-specific problems.
\section{Approach}

In this section, we cover the different components of our study, beginning by describing the policy learning methods we considered in Sec.~\ref{sec:policy-learning}, 14 pre-trained vision models in Sec.~\ref{sec:pre-trained-models}, and 3 simulation environments in Sec.~\ref{sec:3envs}.

\subsection{Policy Learning Methods}
\label{sec:policy-learning}

In general, we consider agents acting within a standard Markov Decision Process (MDP), where at each time step we have access to a tuple $\left(\mathcal{O}, \mathcal{A}, P, R, \gamma \right)$, replete with the usual definitions. The agent for learning motor control consists of an encoder network $f$ and a lightweight controller head $\pi$. The encoder $f$ is a \textit{frozen} pre-trained vision model. Given an image observation $o_t$ from the environment, $f$ first extracts a representation vector $z_t = f(o_t)$. Then, the controller $\pi$ takes in the representation and predicts an action $a_t = \pi(z_t)$. We analyze the following three representative policy learning methods and illustrate them in Figure~\ref{fig:policy_learning_methods}.

\textbf{Reinforcement learning.}
In model-free RL, the goal is to use the in-domain experience to maximize the expected discounted sum of rewards $\mathbb{E}_{\pi}\left[\sum_{t=0}^{\infty} \gamma^t R\left(o_t, a_t\right)\right]$.
 Specifically, we use DrQ-v2~\cite{yarats2021mastering}, a state-of-the-art off-policy actor-critic approach for continuous vision-based control. DrQ-v2 is a representative and also useful choice, given its favorable sample efficiency. Additional algorithmic details are available in Appendix~\ref{app:policy-rl}.

\textbf{Imitation learning through behavior cloning.}
Given access to expert trajectories $\mathcal{T}^e = \{(o^e_t, a^e_t)_{t=0}^T\}_{n=0}^N$, BC corresponds to minimizing the loss function $  \mathbb{E}_{\left(o^e_t, a^e_t\right) \sim \mathcal{T}^e} \left\|a_t^e-\pi(f(o_{t}^e))\right\|_2^2$. Both RL and BC are common approaches to evaluate vision models as frozen perception modules for policy learning in literature. The main role played by the vision model here is to extract representations that contain environment-relevant information. These visual representations are used to replace hand-engineered ground-truth features, which are hard to estimate across diverse control tasks.

\textbf{Imitation learning with a visual reward function.}
Another method to tackle the imitation learning problem involves Inverse Reinforcement Learning (IRL)~\cite{ng2000algorithms}. IRL infers the underlying reward function from the expert trajectories before employing RL to optimize a policy. In our setting, the crucial idea is to craft the underlying reward function based on a distance metric in the vision model's embedding space. We term this the \textit{visual reward function (VRF)}. For example, one straightforward strategy is to define the reward as the negative squared $\ell_2$ distance between the agent's observation and the expert goal image: $-\left\| f(o_t) - f(o_T^e) \right\|_2^2$~\cite{zakka2022xirl}. In our experiments, we adopt a recent algorithm ROT~\cite{haldar2022watch}, which derives the reward based on the Sinkhorn distance~\cite{cuturi2013sinkhorn} between the expert and the agent observations (detailed in Appendix~\ref{app:policy-vrf}). VRF requires not only that the vision model faithfully substitutes ground-truth features, but also that it encodes task progress information in its latent representations. The effectiveness of pre-trained vision models in VRF is underexplored. But such an approach holds great potential since it enables agents to learn directly from diverse human videos~\cite{chen2021learning,kumar2022inverse}.

\subsection{Pre-Trained Vision Models}
\label{sec:pre-trained-models}

\begin{figure*}[!t]
\captionsetup[subfigure]{labelformat=empty}
\captionsetup[subfigure]{aboveskip=2pt,belowskip=3pt}
\centering
\begin{subfigure}{0.19\textwidth}
  \centering
  \includegraphics[width=0.9\linewidth]{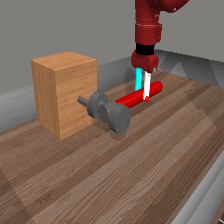}
  \caption{Hammer}
\end{subfigure}
\begin{subfigure}{.19\textwidth}
  \centering
  \includegraphics[width=0.9\linewidth]{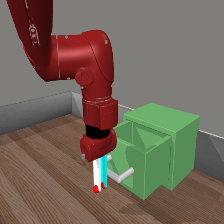}
  \caption{Drawer Close}
\end{subfigure}
\begin{subfigure}{.19\textwidth}
  \centering
  \includegraphics[width=0.9\linewidth]{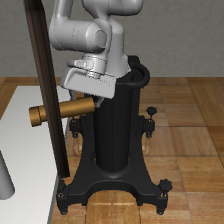}
  \caption{Panda Door}
\end{subfigure}
\begin{subfigure}{.19\textwidth}
  \centering
  \includegraphics[width=0.9\linewidth]{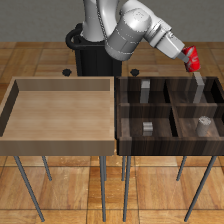}
  \caption{Panda Pick-and-Place}
\end{subfigure}
\begin{subfigure}{.19\textwidth}
  \centering
  \includegraphics[width=0.9\linewidth]{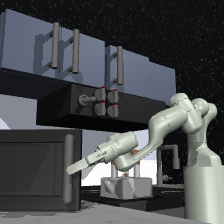}
  \caption{Opening Microwave}
\end{subfigure}
\vskip -0.1in
\caption{Example tasks from 3 environments. Due to the space limit, we only show 2 tasks from Meta-World, 2 tasks from Robosuite, and 1 task from Franka-Kitchen. Please see Appendix~\ref{app:envs} for all 21 tasks.}
\label{fig:env}
\vspace{-1em}
\end{figure*}

\begin{table}[t]
\caption{The highlights of different pre-trained models.}
\label{tab:models}
\vspace{-0.5em}
\resizebox{\columnwidth}{!}{
\begin{small}
\begin{tabular}{lc}
\toprule
Model & Highlights \\
\midrule
  MoCo v2 &   Contrastive learning, momentum encoder\\
  SwAV    &   Contrast online cluster assignments\\
  SimSiam &   Without negative pairs \\
  DenseCL &   Dense contrastive learning, learn local features \\
  PixPro  &   Pixel-level pretext task, learn local features \\
  VICRegL &   Learn global and local features \\
  VFS     &   Encode temporal dynamics \\
  R3M     &   Learn visual representations for robotics \\  
  VIP     &   Learn representations and reward for robotics \\ 
\midrule
 MoCo v3 &  Contrastive learning for ViT \\
 DINO    &  Self-distillation with no labels \\
 MAE     &  Masked image modeling (MIM) \\
 iBOT    &  Combine self-distillation with MIM \\
 CLIP    &  Language-supervised pre-training \\
\bottomrule
\end{tabular}
\end{small}}
\vspace{-1.5em}
\end{table}

We aim to investigate the efficacy of different ``off-the-shelf'' pre-trained vision models for motor control. We consider the following 14 models across 2 architectures. \\
\textbf{ResNet-50}: MoCo v2~\cite{chen2020improved}, SwAV~\cite{caron2020unsupervised}, SimSiam~\cite{chen2021exploring}, DenseCL~\cite{wang2021dense}, PixPro~\cite{xie2021propagate}, VICRegL~\cite{bardes2022vicregl}, VFS~\cite{xu2021rethinking}, R3M~\cite{nair2022r3m} and VIP~\cite{ma2022vip}. \\
\textbf{ViT-B/16}: MoCo v3~\cite{chen2021empirical}, DINO~\cite{caron2021emerging}, MAE~\cite{he2022masked}, iBOT~\cite{zhou2021ibot} and CLIP~\cite{radford2021learning}.

Table~\ref{tab:models} summarizes the highlights of different models, and Appendix~\ref{app:models} holds detailed descriptions. We choose these models for their diversity and coverage of the pre-trained model landscape: \textit{(i)} R3M and VIP are designed with robotic manipulation tasks in mind. \textit{(ii)} DenseCL, PixPro and VICRegL learn local visual features and we speculate that they may benefit motor control, which requires fine-grained spatial information. \textit{(iii)} Other models serve as excellent references given their documentation in previous works, e.g., MoCo v2 and MAE are used in PVR~\cite{parisi2022unsurprising} and MVP~\cite{xiao2022masked}, respectively. 

All models have official open-source codebases, from where we obtain pre-trained weights. One variable that is difficult to control is the pre-training dataset, as it is prohibitively expensive to retrain all the models we considered on the same dataset. However, the pre-training dataset is not a core factor affecting downstream control tasks, as evidenced by Parisi et al.~\yrcite{parisi2022unsurprising}. In addition, all the models are pre-trained on out-of-domain data, i.e., they have never seen a single in-domain image from the environment. Thus, all our subsequent comparisons are on equal footing.

\subsection{Environments}
\label{sec:3envs}

\textbf{Selection criteria.} Our core criteria for selecting benchmark environments is that they are representative of real-world scenarios. An excellent environment should \textit{(i)} support low-level full-physics control (i.e., no magic skills/abstract action space), \textit{(ii)} render visually-realistic observations, and \textit{(iii)} cover diverse tasks and objects. Additionally, dense rewards need to be provided to study RL algorithms, and fast simulation speeds up experimentation. Some environments used in previous works fall short on one or more of our requirements. For example, Habitat~\cite{savva2019habitat} does not support full-physics simulation. DeepMind Control (DMC) Suite~\cite{tassa2018deepmind} focuses on locomotion tasks, and its observations are not visually realistic.

\textbf{Three environments.}
Taking the above factors into consideration, we use a total of 21 tasks across 3 robot manipulation environments: Meta-World (8 tasks), Robosuite (8 tasks), and Franka-Kitchen (5 tasks). Figure~\ref{fig:env} shows sample tasks from each environment. Complete environment details are available in Appendix~\ref{app:envs}. All environments are simulated via the MuJoCo physics engine~\cite{todorov2012mujoco}, which enables fast simulation of physical contact.
We choose distinct challenge tasks from the environments covering various scenes, objects and manipulation skills.

\subsection{Experimental Setup}

\textbf{Implementation details.}
All environment observations are $224 \times 224$ RGB images, which are consistent with the resolution most vision models are pre-trained at. In our experiments, we find that using only one image observation is comparable with using a stack of consecutive images, so we choose the more compute-efficient option. Furthermore, we do not use proprioceptive information (e.g., end-effector poses and joint positions, etc.), ensuring fair comparison of the vision models operating strictly with visual observations. For more details, we refer to Appendix~\ref{app:policy-details}. 

\textbf{Evaluation.}
For each pre-trained vision model and each task, we run 3 seeds of BC and VRF, and 6 seeds of RL (due to its higher variability). After each run, we compute the policy success rate over 100 online rollouts. For the final aggregate performance across tasks, we report interquartile mean (IQM), which is not affected by outliers and has smaller uncertainty (the median and mean scores are shown in Appendix~\ref{app:additioal-exp}). Following the guidelines of Agarwal et al.~\yrcite{agarwal2021deep}, we also report interval estimates via stratified bootstrap confidence intervals (CIs) to further account for uncertainty in results.

\section{Experimental Results}

\begin{figure}[t]
    \centering
    \includegraphics[width=1.0\linewidth]{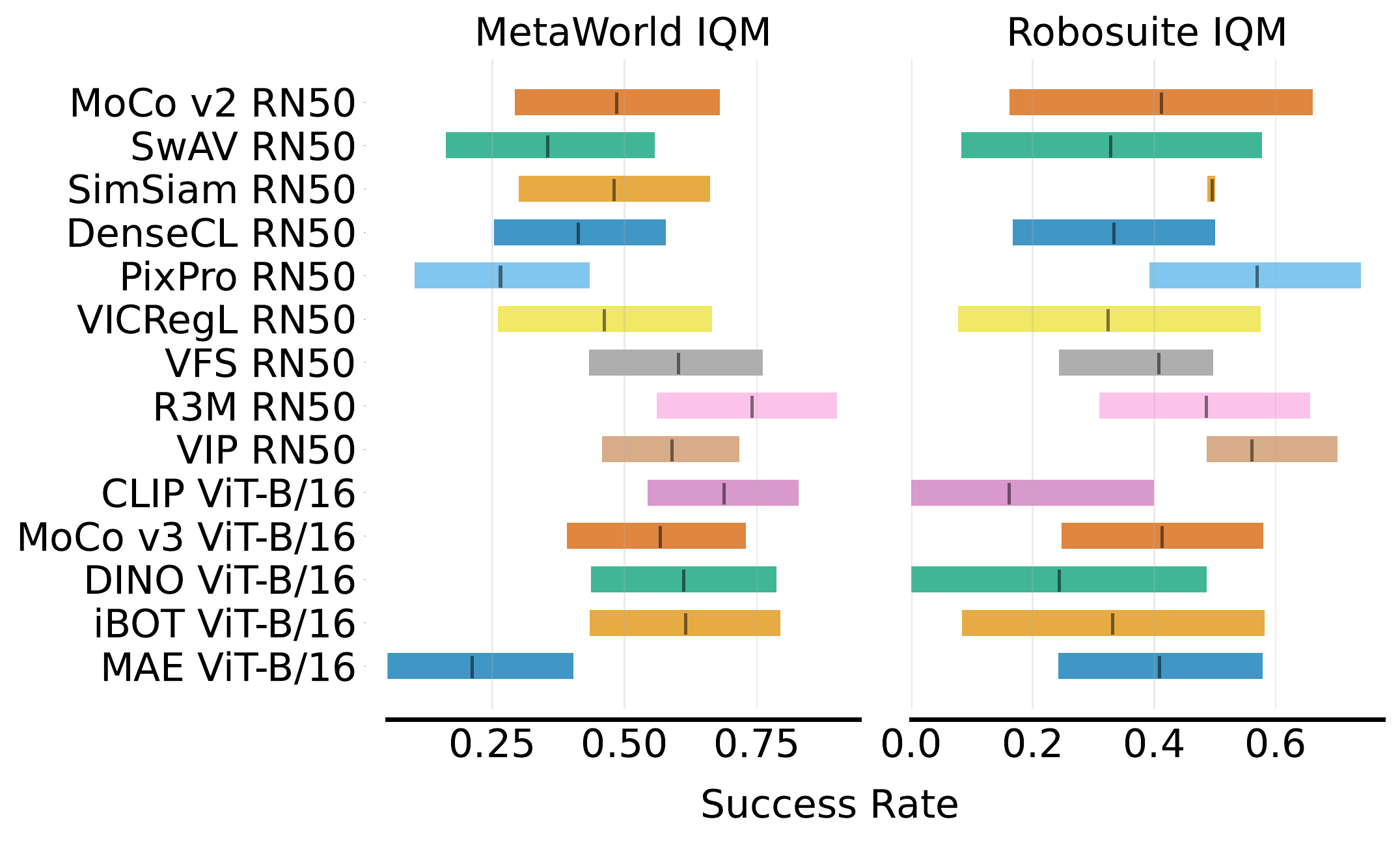} 
    \caption{Aggregate RL performance on Meta-World and Robosuite with 95\% CIs based on 8 tasks per environment. There is no consistency between the two environments.}
    \label{fig:rl-twoenv}
    \vspace{-0.5em}
\end{figure}

\begin{figure}[t]
    \centering
    \includegraphics[width=1.0\linewidth]{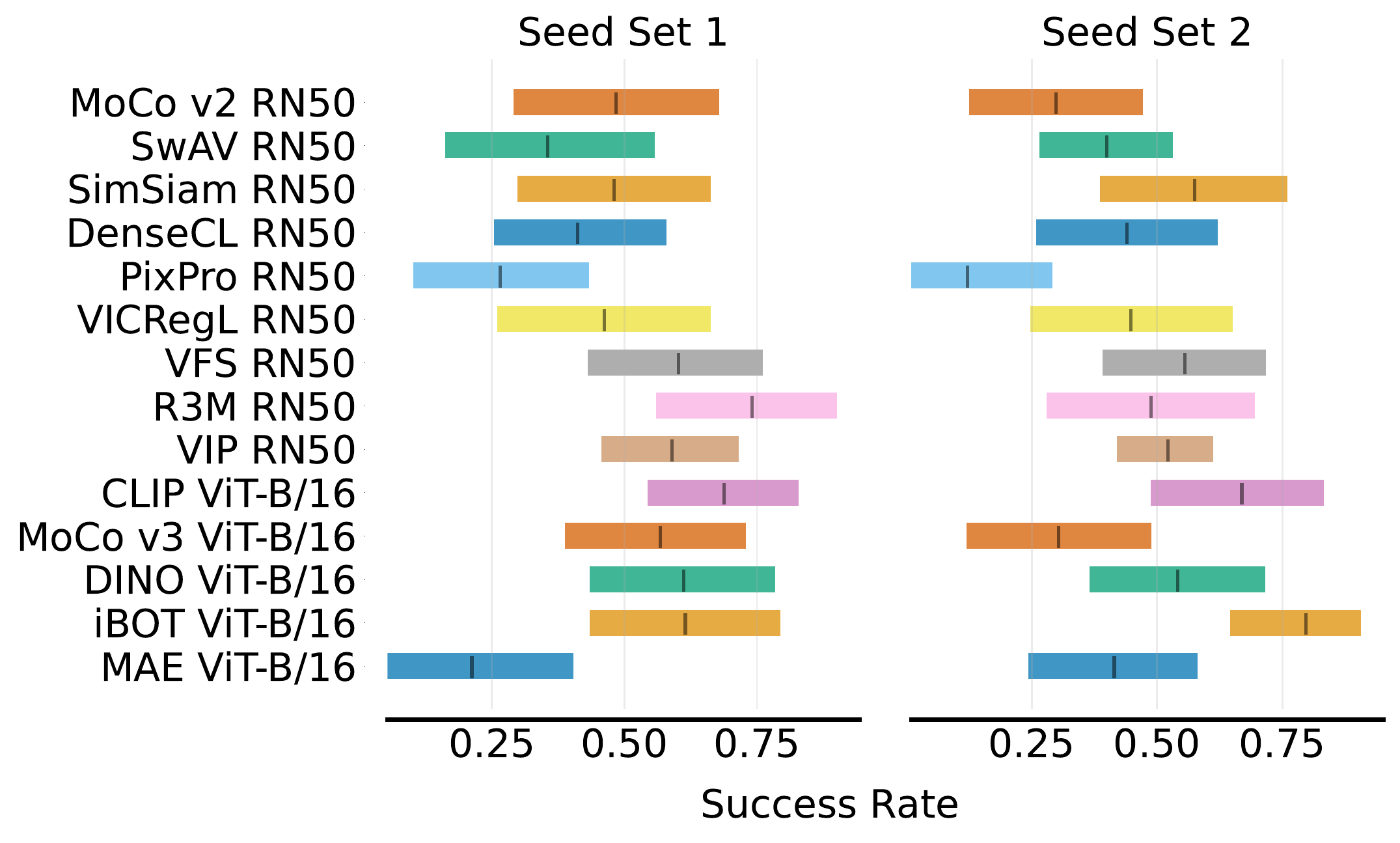} 
    \caption{Aggregate RL performance (IQM scores) of two seed sets on Meta-World. The only difference between the two experiments is the seeds.}
    \label{fig:rl-seeds}
    \vspace{-1em}
\end{figure}

\begin{figure}[t]
    \centering
    \includegraphics[width=1.0\linewidth]{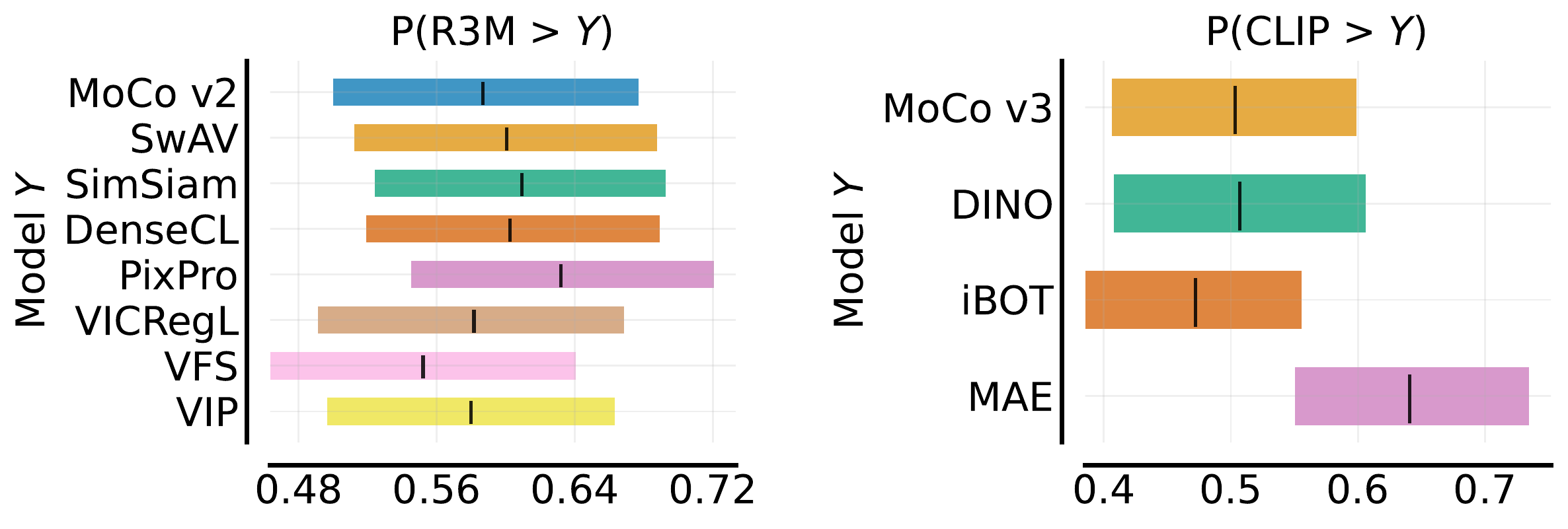} 
    \caption{Probability of improvement on Meta-World with 95\%  CIs. Each subplot shows the probability of improvement of a given vision model compared to all other models.}
    \label{fig:improvement}
    \vskip -1em
\end{figure}

In this section, we first separately analyze the performance of pre-trained vision models for the three policy learning methods. Then, we further investigate what properties of vision models matter for different policy learning methods.

\subsection{Reinforcement Learning}

\textbf{Inconsistency between environments.} 
Figure \ref{fig:rl-twoenv} shows the RL results of all pre-trained vision models on Meta-World and Robosuite. Surprisingly, the rankings of different vision models on the two environments are entirely different. In particular, PixPro ranks second to last on Meta-World but is one of the best-performing models on Robosuite.
The language-supervised CLIP, which performs extremely well on Meta-World, is the worst one on Robosuite. 
At first glance, the effectiveness of a pre-trained vision model seems environment-dependent.

\begin{figure*}[!t]
\captionsetup[subfigure]{labelformat=empty}
\centering
\begin{subfigure}{.95\textwidth}
  \centering
  \caption{ResNet-50}
  \includegraphics[width=\linewidth]{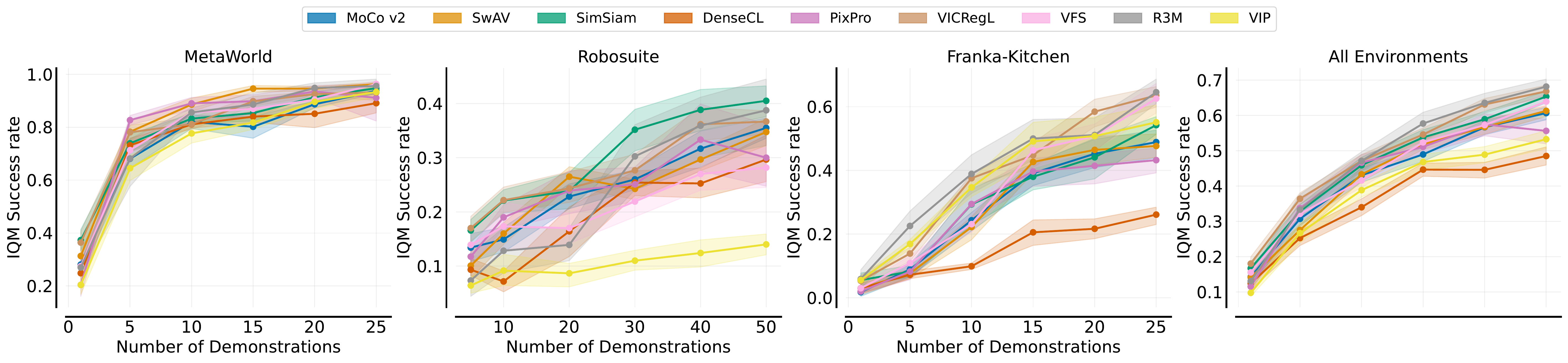}
\end{subfigure} \\
\begin{subfigure}{.95\textwidth}
  \centering
  \caption{ViT-B/16}
  \includegraphics[width=\linewidth]{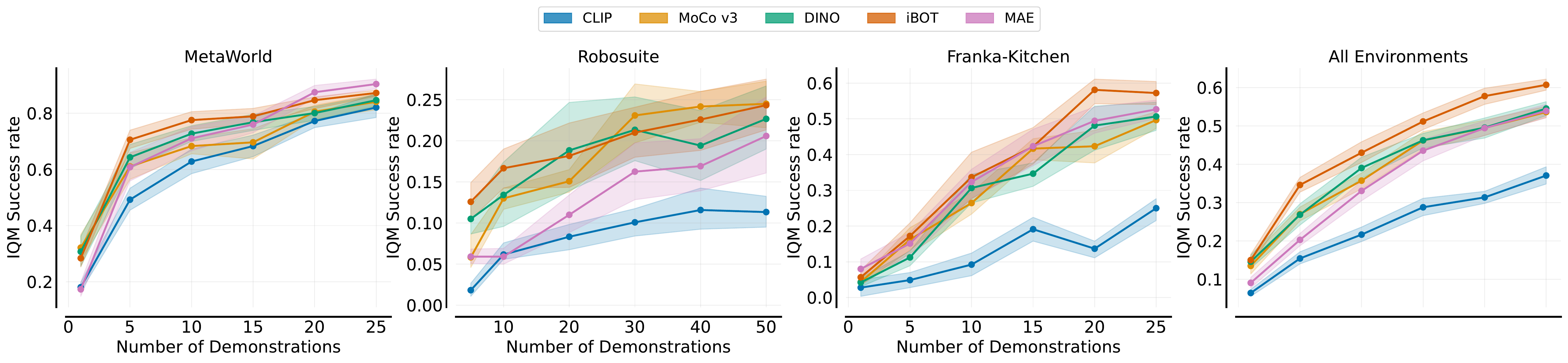}
\end{subfigure}
\caption{BC performance of different pre-trained vision models as a function of the number of demonstrations measured via IQM success rate. Shaded regions show pointwise 95\% percentile stratified bootstrap CIs.}
\label{fig:bc-results}
\vspace{-1em}
\end{figure*}

\textbf{Inconsistency between runs.} 
We further observe that the inconsistency exists not only in different environments, but even in different training runs in the same environment.
Specifically, we repeat the experiment on Meta-World using a different set of 6 seeds. The results are shown in Figure~\ref{fig:rl-seeds}. 
For most models, there is a huge difference in the success rate between the two experiments, with the largest being around 25\% (R3M).
This leads to the fact that even considering only one environment, we cannot reliably conclude which vision model is the best.

\textbf{High variability.}
We hypothesize that the inconsistency between environments and runs is due to the substantial variability of current RL algorithms. This can be confirmed by the 
significantly large confidence intervals (CIs) in Fig.~\ref{fig:rl-twoenv} and Fig.~\ref{fig:rl-seeds}. At the same time, the CIs strikingly overlap for most pre-trained vision models, making it difficult to compare any two models directly.
Hence, we show the \textit{probability of improvement} metric proposed by Agarwal et al.~\yrcite{agarwal2021deep} in Fig.~\ref{fig:improvement}. 
On Meta-World, there is only a 40 - 60\% chance that CLIP improves upon MoCo v3, although CLIP outperforms MoCo v3 by about 20\% in point estimates of IQM (Fig.~\ref{fig:rl-twoenv}, left). We note that the high variability of RL results is also observed in recent works~\cite{henderson2018deep,chan2019measuring,xiao2022masked}. This high variability is due to inherent randomness, which can arise from exploratory choices made during training, stochasticity in the task, and randomly initialized parameters.

\textbf{Not an ideal evaluation method.}
In addition to exhibiting substantial variability, RL is notorious for being sensitive to lower-level choices~\cite{andrychowicz2020matters} like hyperparameter selection. We use the exact same hyperparameters for all vision models, but we find that minor changes to some hyperparameters (e.g., learning rate, buffer size) cause significant differences for a given model. 
All these findings suggest that RL itself is \textit{not} suitable as a downstream policy learning method to evaluate different pre-trained vision models. Further analysis on uncertainty is in Appendix~\ref{app:rl-uncertainty}.

\subsection{Imitation Learning through Behavior Cloning}

We now thoroughly evaluate the large suite of pre-trained vision models using BC.
We train BC policies with varying numbers of expert demonstrations to study how dataset size impacts performance in the low data regime.
The results in Figure~\ref{fig:bc-results} demonstrate that no individual pre-trained vision model can dominate all three environments. However, considering 21 tasks across all environments (Fig.~\ref{fig:bc-results} rightmost), R3M is the best-performing model. This echoes the finding of Nair et al.~\yrcite{nair2022r3m} that R3M enables data-efficient behavior cloning for robotic manipulation. Furthermore, there is a model that follows R3M closely: VICRegL. Interestingly, VICRegL is pre-trained purely on ImageNet~\cite{russakovsky2015imagenet}, without robotic manipulation tasks in mind, while R3M is pre-trained on diverse human videos. This lends credence to a dominant paradigm that advancements in vision can potentially be transferred directly to visuomotor control without extensive control-specific adaptation.

With respect to architectures, iBOT is clearly the best-performing model among ViTs, highlighting the benefit of combining masked image modeling with self-distillation. Another general observation is that ResNet-based models tend to outperform ViT-based models. This is in spite of the fact that model size and computation complexity tend to favor ViTs. We hypothesize this is because transformer-based architectures contain less visual prior knowledge. 

In addition to robust empirical performance, BC is also very simple and sample-efficient. BC shows a high correlation with linear probing results (see Sec.~\ref{sec:deeper-understanding}), which can reflect whether the visual features encode environment-related information. These all verify the reliability of using BC to evaluate different pre-trained vision models.

\subsection{Imitation Learning with Visual Reward Functions.}

\begin{figure*}[t!]
    \centering
    \includegraphics[width=1.0\linewidth]{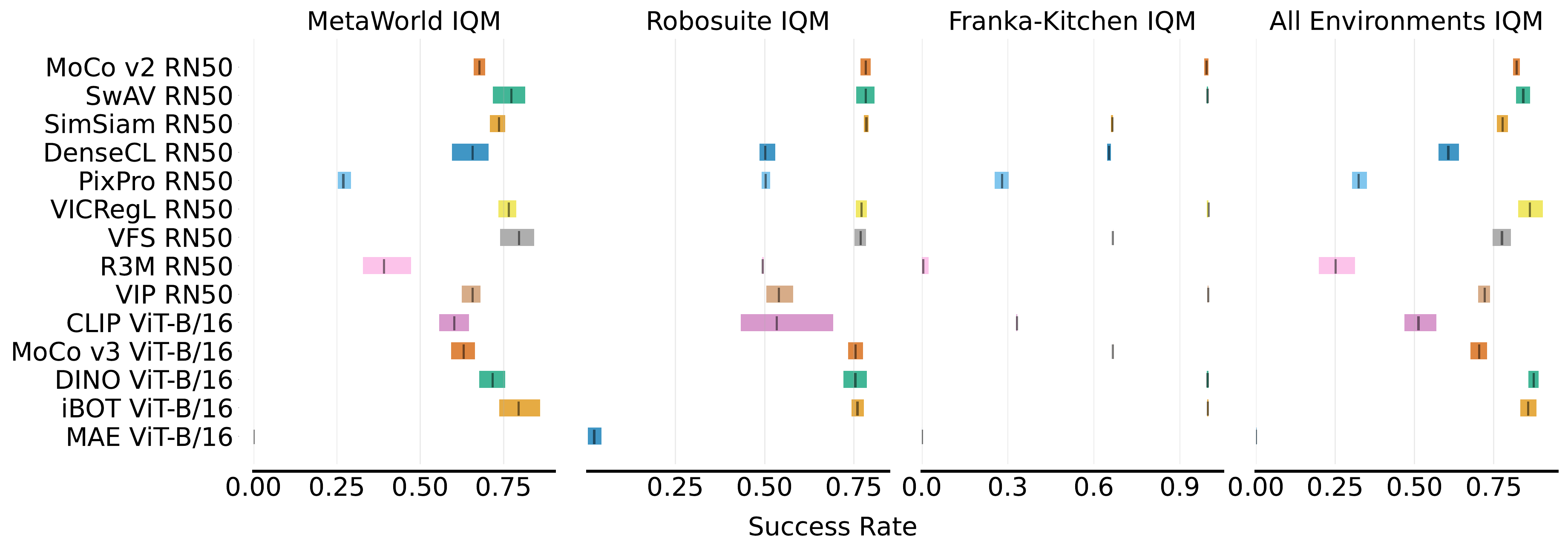}
    \vspace{-2em}
    \caption{Aggregate VRF performance (IQM scores) of different pre-trained vision models with 95\% CIs. The models rely on joint embedding architectures perform well. The results of MAE are close to \textit{zero}.}
    \label{fig:rot-results}
    \vspace{-1em}
\end{figure*}

Figure~\ref{fig:rot-results} shows the VRF results of different pre-trained vision models on three environments. There are several intriguing observations:
\textit{(i)} All performant models across the board (e.g., MoCo v2, VICRegL, DINO, etc.) share one commonality: they are based on \textit{joint embedding architectures}~\cite{lecun2022path} that force the global features to be invariant to a sampling process selecting pairs of different views of the same image.
\textit{(ii)} Perhaps most surprisingly, MAE exhibits near \textit{zero} performance on all three environments, which means that the rewards computed from MAE representations are entirely meaningless (detailed analysis in the next section).
\textit{(iii)} R3M, the star model on BC, only obtains 25\% IQM success rate when considering all environments tasks, ranking as the second-worst model.
\textit{(iv)} Some models that emphasize learning local image characteristics (e.g., PixPro and DenseCL) also do not perform well.

Summarising these results, we see that the ideal pre-trained vision model that VRF requires is starkly different from the one that RL and BC need. Specifically, an excellent vision model for VRF should learn features at a global scale.
Additionally, the invariance properties brought by joint embedding architectures outweigh the amount and type of pre-training data.

Notably, different pre-trained vision models yield very consistent performance when using VRF.
Moreover, VRF can reveal whether the visual representations capture a notion of task progress. 
Finally, the policies trained with VRF use only one expert demonstration and achieve high performance. All these merits make VRF a strong candidate for evaluating different vision models for motor control.

\subsection{Understanding Properties of Vision Models}
\label{sec:deeper-understanding}

The prior experiments focus on performance-driven comparisons of pre-trained vision models using different policy learning methods. But what specific properties of various vision models enable their comparative advantage?

\begin{figure}[t]
    \centering
    \includegraphics[width=0.9\linewidth]{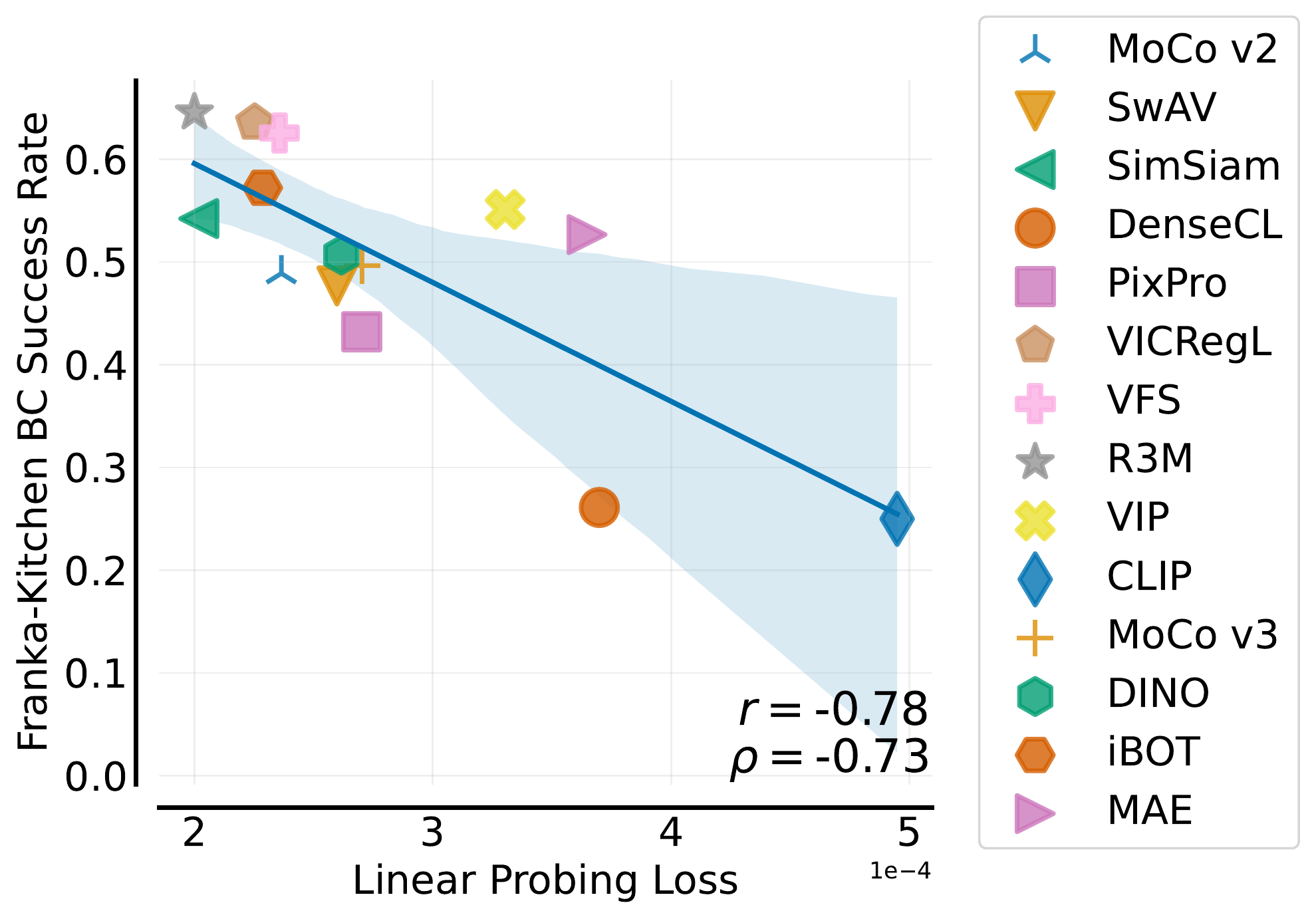} 
    \caption{Correlation between Franka-Kitchen linear probing loss on the validation set and BC success rate (IQM scores) based on 5 tasks. Correlation coefficients (Pearson's $r$ and Spearman's $\rho$) are shown in the bottom right.}
    \label{fig:probing-bc-corr}
    \vspace{-1.5em}
\end{figure}

\textbf{Properties crucial for BC.}
The first question we set out to answer is what information is necessary for data-efficient behavior cloning.
We borrow the commonly employed linear probing protocol in self-supervised visual representation learning~\cite{zhang2016colorful,oord2018representation} to facilitate our analysis.
Specifically, we train a single fully-connected layer on top of the frozen pre-trained features to predict the hand-engineered environment states. We train this linear regressor using images from the deployment environment by minimizing the mean squared error between the predictions and ground-truth state features.
We use the validation loss as a proxy for the quality of visual features. Lower loss indicates that more environment-relevant information, such as object locations and joint positions, is retained in the pre-trained visual representations.
Details are in Appendix~\ref{app:probing-details}.

For Franka-Kitchen, we show the correlation coefficients between linear probing loss and BC success rate in Figure~\ref{fig:probing-bc-corr}. We observe a strong \textit{inverse correlation}, with Pearson's $r$ being $-0.78$ and Spearman's $\rho$ being $-0.73$. 
This indicates that pre-trained vision models that effectively encode ground-truth environment information will lead to more capable BC agents. Our findings also suggest that the linear probing protocol can be a valuable and intuitive alternative for evaluating vision models for motor control.

\begin{figure}[t]
    \centering
    \includegraphics[width=1.0\linewidth]{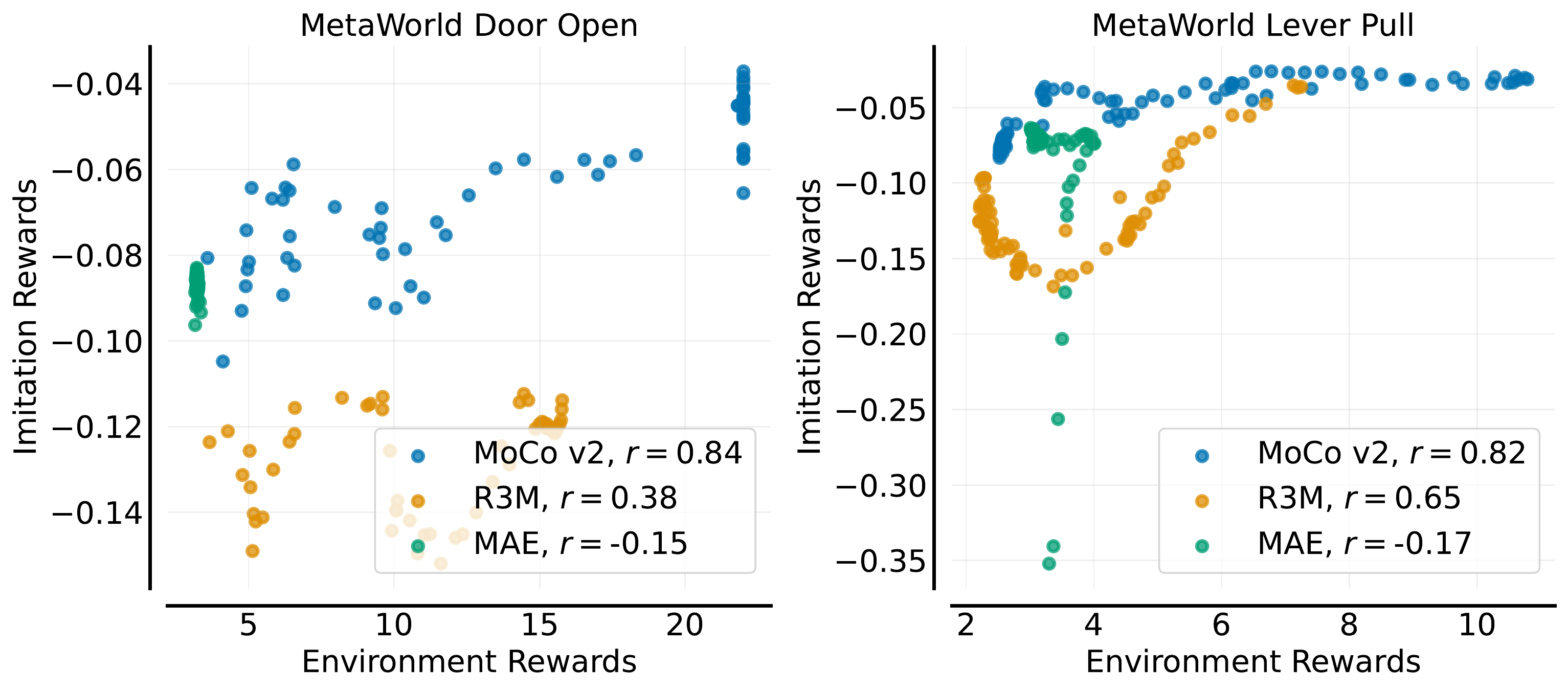} 
    \caption{Environment rewards (x-axes) vs. imitation rewards correlation (y-axes).
    We plot the rewards collected using one online rollout after training.
    }
    \label{fig:rewards-corr}
    \vspace{-4mm}
\end{figure}

\textbf{Properties crucial for VRF.}
We next explore what causes considerable distinctions in the VRF performance of different models.
We first examine the relationship between hand-designed environment rewards and imitation rewards computed in the vision model's embedding space. The environment rewards capture human intuition for how a task should be solved. The scatterplots of three typical models on two Meta-World tasks are shown in Figure~\ref{fig:rewards-corr}. The results show that the imitation rewards of MoCo v2 exhibit a much stronger correlation with the environment rewards, while there is little to no correlation for MAE. R3M falls somewhere in between. The uninformative rewards cause MAE to have near \textit{zero} VRF scores on all control tasks.

\begin{figure}[t]
\captionsetup[subfigure]{labelformat=empty}
\centering
\begin{subfigure}{.3\columnwidth}
  \centering
  \includegraphics[width=\textwidth]{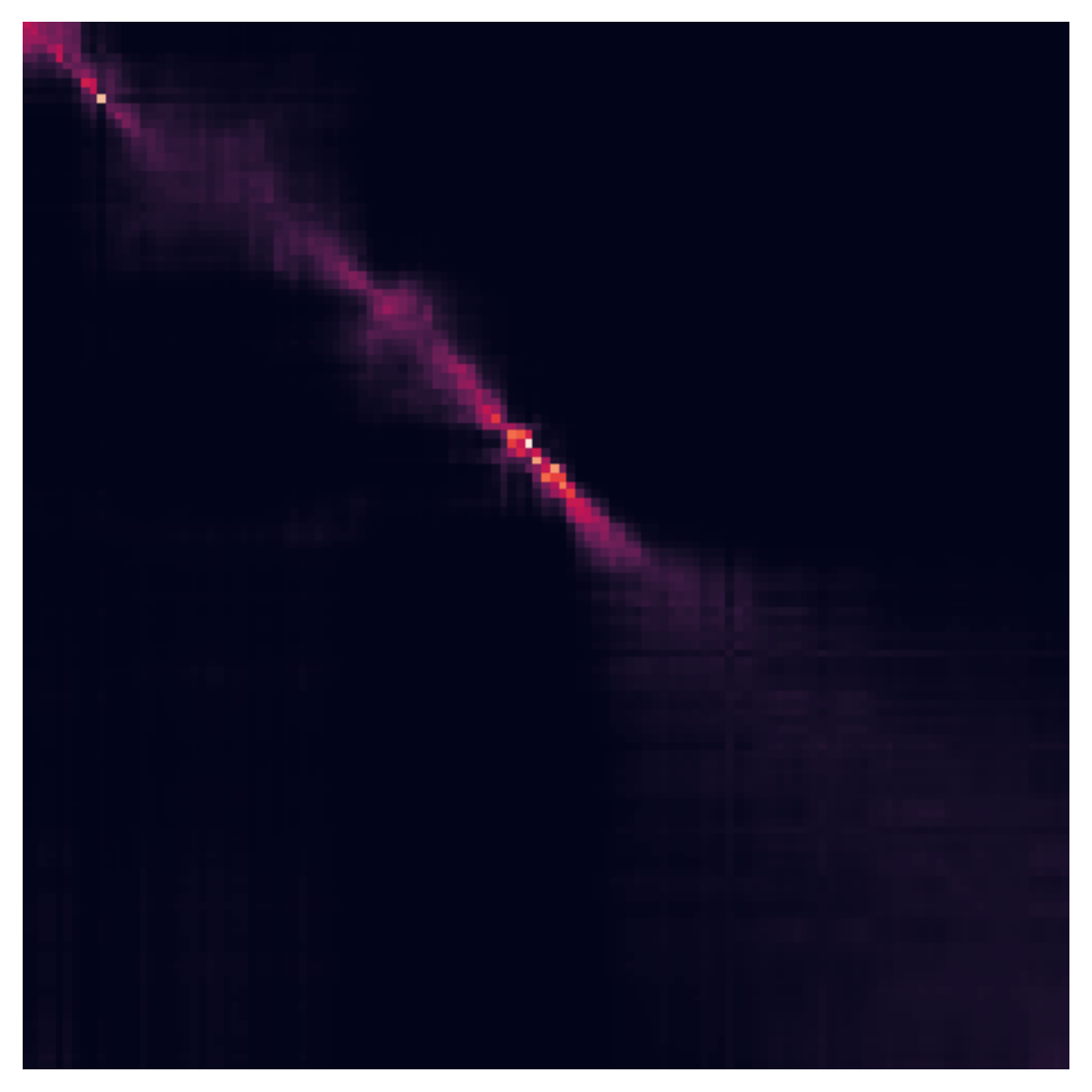}
    \caption{MoCo v2}
\end{subfigure} 
\begin{subfigure}{.3\columnwidth}
  \centering
  \includegraphics[width=\textwidth]{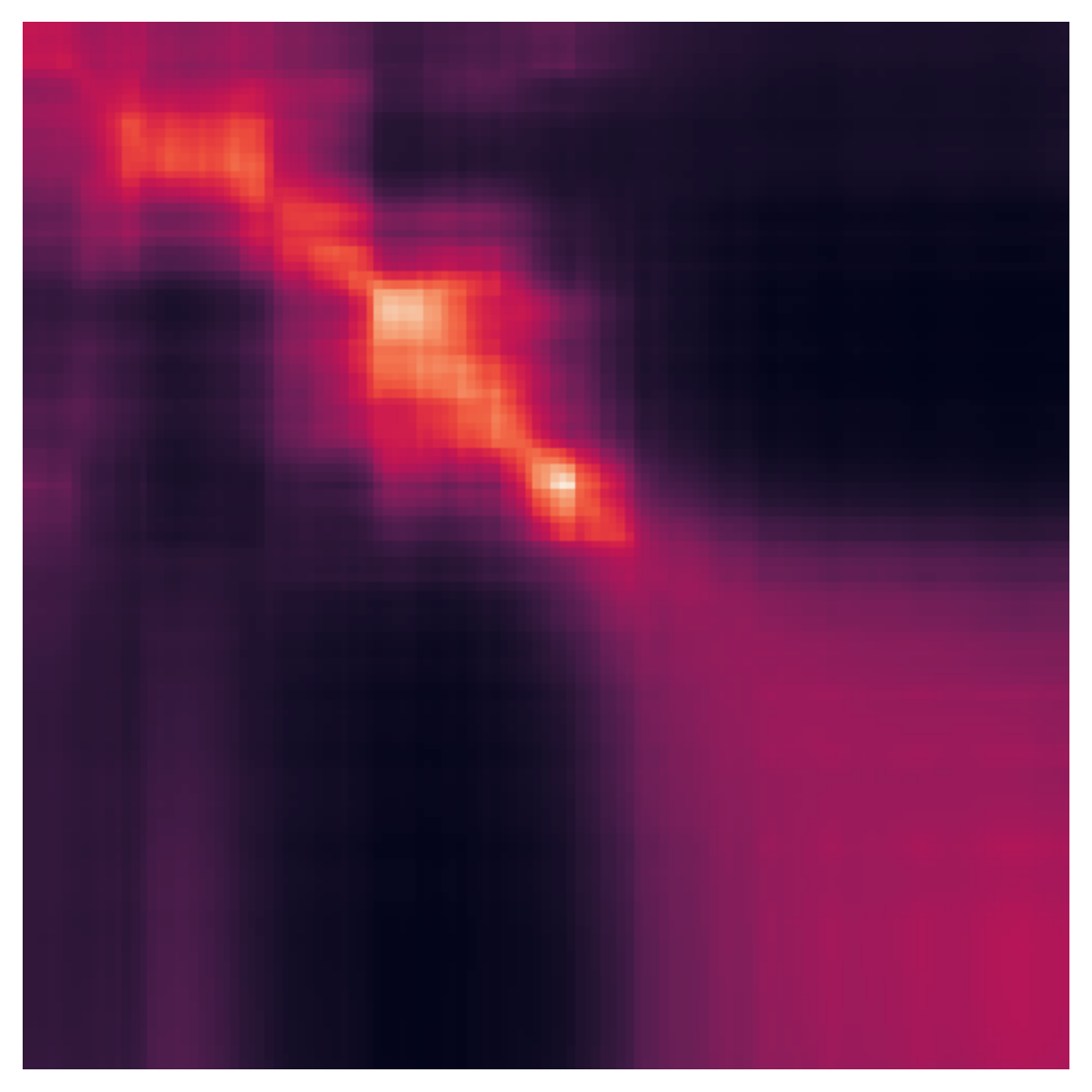}
  \caption{R3M}
\end{subfigure}
\begin{subfigure}{.3\columnwidth}
  \centering
  \includegraphics[width=\textwidth]{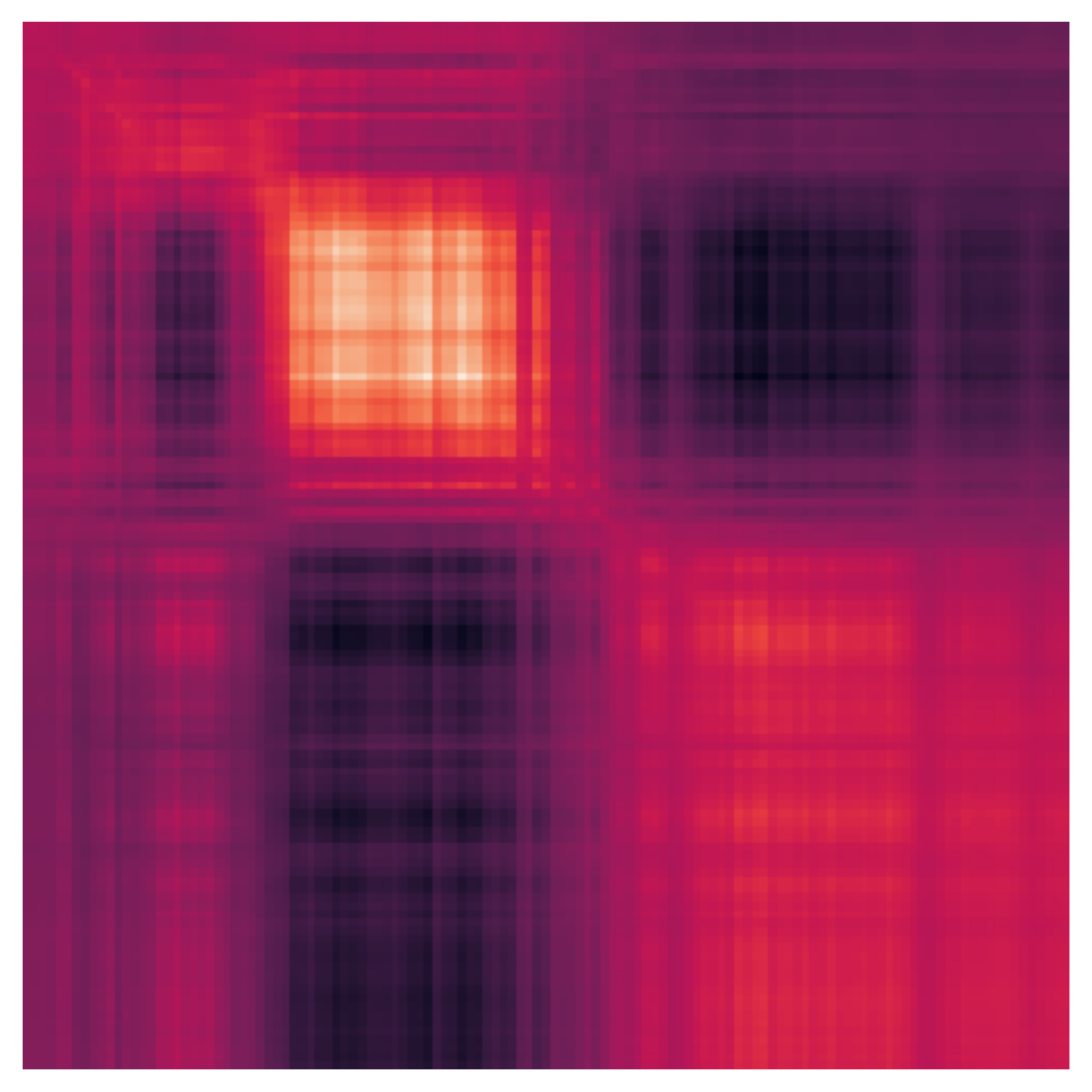}
  \caption{MAE}
\end{subfigure}
\caption{OT matrix between two expert trajectories on Meta-World \texttt{door-open}. Brighter colors indicate higher values. This OT matrix is obtained by solving Equation~\ref{eq:ot}.}
\label{fig:ot-vis}
\vspace{-1em}
\end{figure}

\begin{figure}[t]
    \centering
    \includegraphics[width=0.85\linewidth]{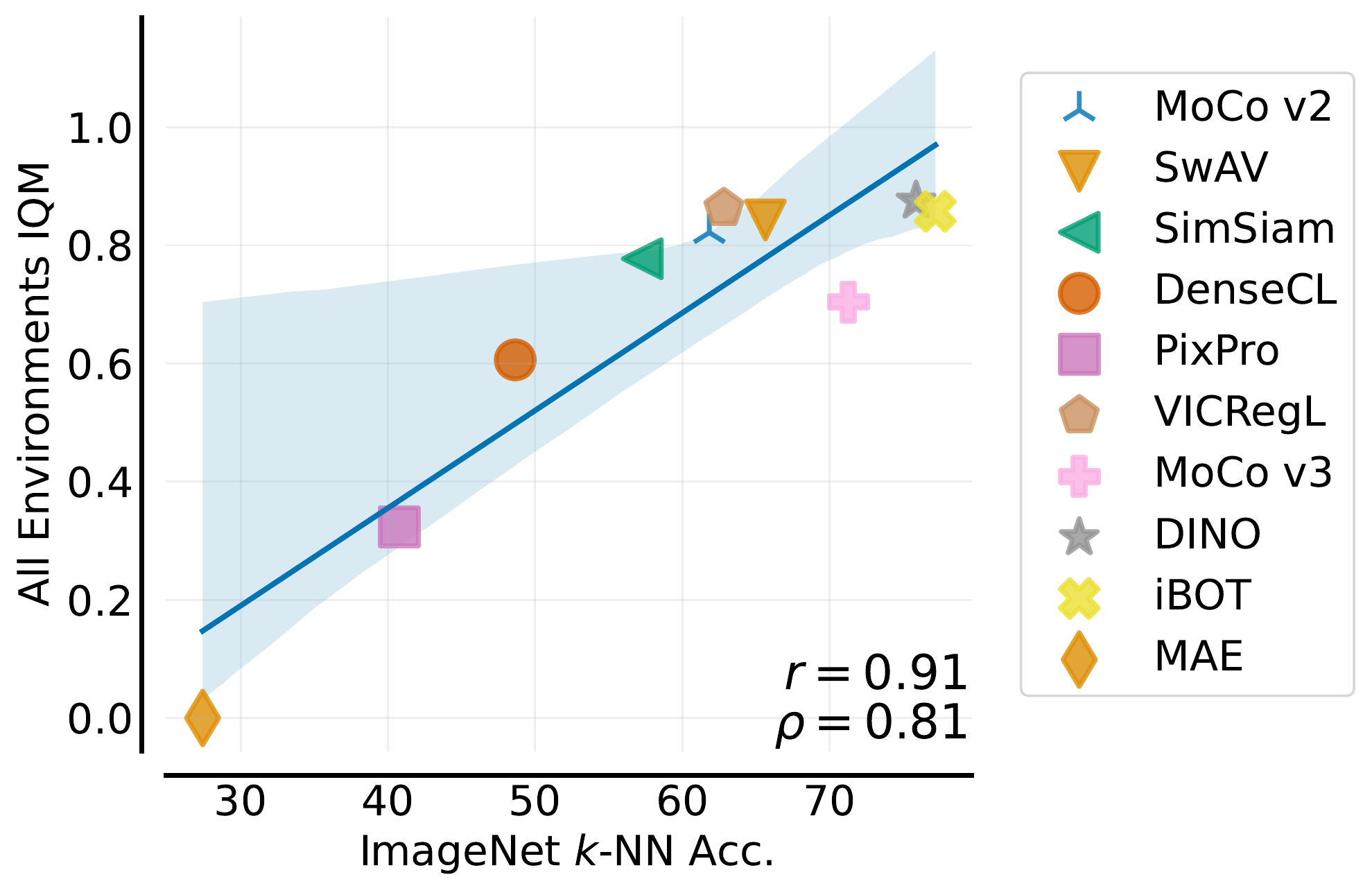} 
    \caption{Correlation between ImageNet $k$-NN classification accuracy and VRF performance on all environments. We show all the vision models pre-trained on ImageNet.}
    \label{fig:knn-rot-corr}
    \vspace{-1.5em}
\end{figure}

To gain a deeper understanding of the reward correlation differences among vision models, we visualize the optimal transport (OT) matrix in the ROT algorithm. The OT matrix illustrates the correspondences/optimal alignment between two distributions. Specifically, we compute the OT between two expert trajectories. Since the two expert trajectories exhibit similar behaviors, the OT matrix computed from an ideal vision model should have high values on the approximate \textit{diagonal}. The results are shown in Figure~\ref{fig:ot-vis}. We observe that the high values of MoCo v2 are concentrated on the diagonal, while the values of MAE are not, indicating that for a given image observation, nearly all the images in the other trajectory are equally similar. This suggests that MAE features cannot capture a notion of task progress.

We further find a metric that is highly predictive of VRF performance: ImageNet $k$-NN classification accuracy, as shown in Figure~\ref{fig:knn-rot-corr}. The correlation between $k$-NN accuracy and VRF performance aggregated on all environments is as high as $r=0.91$. Note that the poor performance of MAE on both VRF (near \textit{zero}) and $k$-NN classification (27.4\%) highlights the difficulty of using MAE features directly through simple similarity metrics such as cosine similarity. 
We hypothesize that there is a shared challenge in using frozen features directly for the ``masked signal modeling'' pre-training paradigm. 
Some previous works in NLP attribute this to the anisotropic problem~\cite{ethayarajh2019contextual,li2020sentence}: the representations are not uniformly distributed with respect to direction and only occupy a narrow cone in the embedding space. To achieve good VRF performance, we expect visual representations can be utilized directly without further compute-intensive fine-tuning.

\section{Conclusion and Discussion}

The proliferation of pre-trained vision models for visuomotor control is an exciting and ongoing event. 
However, there has been a noticeable lack of work on evaluation protocols in this area. We conduct the first thorough empirical evaluation of pre-trained vision model performance across different downstream policy learning methods and environments.

Our evaluation shows the following insights: (\textit{i}) The effectiveness of a pre-trained vision model highly depends on the downstream policy learning methods. The vision of a `\textit{universal}' pre-trained model with the best performance on all control tasks is yet to be realized. (\textit{ii}) Due to randomness that cannot be mitigated, RL methods demonstrate overly high variability, and thus cannot be positioned as reliable evaluation methods for vision models. (\textit{iii}) Without control-specific adaptation, BC still benefits from the latest benchmark-leading models in the vision community, due to their inherent ability to capture more environment-relevant information such as object locations and joint positions. (\textit{iv}) Different vision models yield the most consistent performance when using VRF, which requires the vision model to learn global features and capture a notion of task progress. MAE is a noticeable underperforming outlier, likely due to the fact that it suffers from the anisotropic problem.

Our findings highlight the importance of evaluating vision models for motor control in a comprehensive manner. We advocate for using more consistent and robust evaluation methods such as VRF and BC to minimize uncertainty and obtain reliable results, as well as claim the effectiveness of the pre-trained vision models in terms of which specific downstream policy learning algorithms are targeted. Although we are unable to provide guidance to researchers who rely on RL due to the unavailability of expert demonstrations, we believe that this limitation is temporary. Our work, as well as other related studies, demonstrate the high variability of RL, which could motivate researchers to develop low-variance RL algorithms. Once such an algorithm is available, it will be possible to evaluate different pre-trained vision models using RL.

With the rapid development of pre-trained vision models for motor control, we believe there is an urgent need to benchmark their empirical performance.
To further support researchers in this endeavor, we will release a library including our evaluation benchmark and pre-trained models. We hope our work will help measure progress in this field and provide common ground for future comparisons.

\section*{Acknowledgements}
This work is supported by the Ministry of Science and Technology of the People's Republic of China, the 2030 Innovation Megaprojects ``Program on New Generation Artificial Intelligence'' (Grant No. 2021AAA0150000).
This work is also supported by the National Key R\&D Program of China (2022ZD0161700).
\bibliography{main}
\bibliographystyle{icml2023}

\cleardoublepage
\appendix
\counterwithin{figure}{section}
\counterwithin{table}{section}
\onecolumn

\section{Policy Learning Methods}
\label{app:policy}

\subsection{Reinforcement Learning: DrQ-v2}
\label{app:policy-rl}
We employ DrQ-v2~\cite{yarats2021mastering}, an off-policy actor-critic approach for continuous vision-based control, as our reinforcement learning algorithm. The original formulation of DrQ-v2 relies on data augmentations (e.g., random shifts)~\cite{kostrikov2020image,laskin2020reinforcement} to facilitate the learning of a shallow \textit{randomly initialized} image encoder. Our \textit{frozen} encoder network $f$ is pre-trained on real-world images, and the representations already possess some general knowledge about the environment. Thus, we do not apply any data augmentation to the image observation. This makes it possible to store representation vectors $z_t=f\left(o_t\right)$ directly in the replay buffer instead of raw image observations. The actor and critic networks can take in the representations sampled from the replay buffer, without re-encoding via the encoder, significantly speeding up the training process.

The core of DrQ-v2 is Deep Deterministic Policy Gradient (DDPG)~\cite{lillicrap2015continuous} augmented with $n$-step returns.  
The critic is trained using clipped double Q-learning~\cite{fujimoto2018addressing} to reduce the overestimation bias in the target value. Specifically, this requires training two Q-functions $Q_{\theta_{1}}$ and $Q_{\theta_{2}}$. The critic loss for each Q-function is given by:

\begin{equation}
    \mathcal{L}_{\theta_k}(\mathcal{D})=\mathbb{E}_{\tau \sim \mathcal{D}}\left[\left(Q_{\theta_k}\left(z_t, a_t\right)-y\right)^2\right] \quad \forall k \in\{1,2\}
\end{equation}

where $\tau$ denotes a mini-batch of transitions $\left(z_t, a_t, r_{t: t+n-1}, z_{t+n}\right)$ sampled from the replay buffer $\mathcal{D}$ and $y$ is the TD target defined as:

\begin{equation}
y=\sum_{i=0}^{n-1} \gamma^i r_{t+i}+\gamma^n \min _{k=1,2} Q_{\bar{\theta}_k}\left(z_{t+n}, a_{t+n}\right)
\end{equation}

Here, $a_{t+n}=\pi_\phi\left(z_{t+n}\right)+\epsilon$ and $\bar{\theta}_1, \bar{\theta}_2$ are the slow-moving weights of target Q-networks. The exploration noise $\epsilon$ is sampled from $\operatorname{clip}\left(\mathcal{N}\left(0, \sigma^2\right),-c, c\right)$, with variance $\sigma^2$ following a linear decay schedule.
Finally, the deterministic actor $\pi_\phi$ is trained using deterministic policy gradients (DPG)~\cite{silver2014deterministic} with the following loss:

\begin{equation}
\mathcal{L}_\phi(\mathcal{D})=-\mathbb{E}_{z_t \sim \mathcal{D}}\left[\min _{k=1,2} Q_{\theta_k}\left(z_t, a_t\right)\right]
\end{equation}

where $a_t = \pi_\phi(z_t) + \epsilon$, and $\epsilon \sim \operatorname{clip}\left(\mathcal{N}\left(0, \sigma^2\right),-c, c\right)$.

\subsection{Imitation Learning with Visual Reward Functions: ROT}
\label{app:policy-vrf}

We adopt Regularized Optimal Transport (ROT)~\cite{haldar2022watch} as one representative algorithm for imitation learning with a visual reward function. ROT derives the imitation reward $r^e$ based on the Sinkhorn distance between the agent and the expert observations. In particular, ROT interprets an observation trajectory $\left(o_1, \ldots, o_T\right)$ as a discrete probability measure of the form $\mu_{o}=\frac{1}{T} \sum_{t=1}^T \delta_{o_t}$. The closeness between agent trajectories $o^a_{1:T}$ and expert trajectories $o^e_{1:T}$ can be computed by measuring the optimal transport of probability mass from $o^a_{1:T} \rightarrow o^e_{1:T}$. We encode trajectories using \textit{frozen} pre-trained vision models:

\begin{equation}
z^a=\left[f\left(o_1^a\right), \ldots, f\left(o_T^a\right)\right] \quad
z^e=\left[f\left(o_1^e\right), \ldots, f\left(o_T^e\right)\right]
\end{equation}

Given a cosine cost matrix computed between encoded image observations $C_{t, t^{\prime}}= 1 - \frac{\langle z^a_t, z^e_{t^{\prime}}\rangle}{\left\|z^a_t\right\| \cdot \left\|z^e_{t^{\prime}}\right\|}$, the optimal alignment between an encoded expert trajectory $z^e$ and an encoded agent trajectory $z^a$ can be computed as:

\begin{equation}
\label{eq:ot}
    \mu^* \in \operatorname*{argmin}_{\mu \in \mathcal{M}} \sum_{t, t^{\prime}=1}^T C_{t, t^{\prime}} \mu_{t, t^{\prime}}
\end{equation}

where $\mathcal{M}=\left\{\mu \in \mathbb{R}^{T \times T}: \mu \mathbf{1}=\mu^T \mathbf{1}=\frac{1}{T} \mathbf{1}\right\}$ is the set of coupling matrices. Since solving Eq.~\ref{eq:ot} is computationally expensive, ROT uses the Sinkhorn algorithm to get approximate solutions quickly~\cite{cuturi2013sinkhorn,papagiannis2020imitation,cohen2021imitation}.

Finally, the imitation reward for each agent observation can be extracted using the equation:

\begin{equation}
\label{eq:reward}
r^e\left(o_t^a\right)=-\sum_{t^{\prime}=1}^T C_{t, t^{\prime}} \mu_{t, t^{\prime}}^*
\end{equation}

Given the computed imitation reward, ROT employs DrQ-v2 as the underlying RL optimizer to achieve efficient reward maximization. Intuitively, this encourages the agent to generate trajectories that closely match demonstrated expert trajectories.

ROT combines the inverse reinforcement learning (IRL) process with behavior cloning (BC) to further improve sample efficiency and final performance. This is done in two phases. In the first phase, BC is used to pre-trained a randomly initialized policy. In the second phase, the pre-trained policy is fine-tuned with the IRL objective and BC objective simultaneously, where the BC loss is added to the IRL objective with an adaptive weight. Strictly speaking, ROT is not a pure IRL algorithm. However, the primary role of BC is to `regularize' IRL, and we empirically find that the quality of the imitation reward is the decisive factor. This is evidenced by the poor performance of MAE (see Figure~\ref{fig:rot-results}). Despite the assistance of BC, the scores achieved by MAE on most tasks are still \textit{zero}. We believe that using ROT as a downstream policy learning method can reflect the suitability of different pre-trained vision models as good visual reward functions.

\section{Pre-Trained Vision Models}
\label{app:models}

Throughout the paper, we consider 14 pre-trained vision models covering prevalent pre-training methods. Here, we give a brief description of each of the models.

\textbf{MoCo v2}~\cite{he2020momentum,chen2020improved} is a classic unsupervised visual representation learning method. 
MoCo v2 builds upon the instance discrimination task~\cite{wu2018unsupervised} that considers each image of the dataset (or “instance”) and its transformations as a separate class. In addition, MoCo v2 uses an explicit momentum encoder to build large and consistent negative samples for contrastive learning.
The representations learned transfer well and match the performance of the supervised pre-training counterpart. 
PVR~\cite{parisi2022unsurprising} finds that MoCo v2 representations can be competitive or even better than hand-engineered ground-truth features to train motor control policies. We note, that in contrast to PVR, we do not use representations from multiple layers and only use the representations from $\texttt{res5}$ block (last block) for fair comparisons with other ResNet models.

\textbf{SwAV}~\cite{caron2020unsupervised} is a clustering-based method~\cite{caron2018deep} for unsupervised visual representation learning. SwAV forces the representations of different images to belong to different clusters on the unit sphere, which is achieved by computing the assignment from one image view and predicting it from another image view. SwAV performs online clustering under a balanced partition constraint for each batch, which ensures that the assignment to clusters is as uniform as possible.

\textbf{SimSiam}~\cite{chen2021exploring} uses simple Siamese networks~\cite{bromley1993signature} to learn meaningful representations by directly maximizing the similarity of one image’s two views. Neither negative pairs nor a momentum encoder is used. The stop-gradient operation in SimSiam plays an essential role in avoiding collapsing solutions.

\textbf{DenseCL}~\cite{wang2021dense}, short for dense contrastive learning, is a self-supervised learning method that operates directly at the levels of local features. DenseCL defines the positive sample of each local feature vector by extracting the correspondence across two views and optimizing a pairwise contrastive (dis)similarity loss at the local feature level. The learned representations preserve spatial information, which is beneficial for dense prediction tasks like semantic segmentation and object detection.

\textbf{PixPro}~\cite{xie2021propagate} is similar to DenseCL in that it utilizes dense pretext tasks for self-supervised visual representation learning. But PixPro uses a BYOL-style (i.e., non-contrastive)~\cite{grill2020bootstrap} training framework, eliminating the need for negative samples. Specifically, PixPro learns the representations by pulling local feature vectors belonging to the same spatial region close together.

\textbf{VICRegL}~\cite{bardes2022vicregl} combines the best of global feature learning and local feature learning. This is achieved by applying the VICReg criterion~\cite{bardes2021vicreg} to pairs of global feature vectors and pairs of local feature vectors simultaneously. Two local feature vectors are designated as positive pairs and attracted to each other if their $\ell_2$ distance is below a threshold or if their relative locations are consistent with a known geometric transformation between the two input views. VICRegL achieves excellent performance on detection and segmentation tasks while maintaining good performance on classification tasks.

\textbf{VFS}~\cite{xu2021rethinking} is self-supervised pre-trained on a large-scale video dataset, Kinetics400~\cite{kay2017kinetics}, while all the other models above are pre-trained on a static image dataset, ImageNet~\cite{russakovsky2015imagenet}. The pre-training pipeline of VFS is similar to that of SimSiam, with the exception that VFS considers frames at different video timestamps as different views for similarity learning. We use VFS to investigate if representations that capture temporal information are beneficial for control policy learning.

\textbf{R3M}~\cite{nair2022r3m} is a vision model designed to enable data-efficient behavior cloning for robotic manipulation tasks. R3M is pre-trained on the large-scale Ego4D human video dataset~\cite{grauman2022ego4d}. R3M leverages time-contrastive learning~\cite{sermanet2018time}, video-language alignment, and an L1 penalty to encourage sparse and compact representations. R3M represents a significant advancement in the field of pre-trained vision models for motor control, and should be taken into account in our large-scale benchmarking studies.

\textbf{VIP}~\cite{ma2022vip} is designed to provide visual representations and dense reward signals for robotic manipulation tasks. VIP treats representation learning as an offline goal-conditioned reinforcement learning problem and solves the Fenchel dual problem of goal-conditioned value function learning that does not depend on actions, enabling pre-training on large-scale human video datasets.

\textbf{MoCo v3}~\cite{chen2021empirical} is a straightforward extension of MoCo v2 by replacing the backbone architecture from ResNet~\cite{he2016deep} to Vision Transformers (ViT)~\cite{dosovitskiy2020image}. However, instability is a major issue in self-supervised ViT training. MoCo v3 alleviates the instability issue by simply freezing the patch projection layer in ViT.

\textbf{DINO}~\cite{caron2021emerging} is a simple approach for self-supervised ViT training that can be interpreted as a form of knowledge distillation~\cite{hinton2015distilling} without labels. Specifically, DINO uses a student network to predict the output of a momentum-updated teacher network with a standard cross-entropy loss. To avoid collapse solutions, a centering and sharpening operation is applied to the momentum teacher outputs. The resulting features of DINO are biased towards shape and explicitly contain the scene layout information of an image.

\textbf{MAE}~\cite{he2022masked} is pre-trained by the masked image modeling (MIM) task~\cite{bao2021beit} inspired by the success of masked language modeling in NLP~\cite{devlin2018bert}. MAE adopts an asymmetric encoder-decoder architecture. The encoder is only applied to the visible subset of patches (without mask tokens). The lightweight decoder reconstructs the image from the encoded visible patches and mask tokens. The mask ratio is as high as 75\% to reduce the heavy spatial redundancy of images. MVP~\cite{xiao2022masked} uses an MAE model pre-trained on images in the wild, e.g., from YouTube~\cite{shan2020understanding} or Egocentric videos~\cite{damen2018scaling} to provide effective representations for motor control. In this study, we adopt the original MAE pre-trained on ImageNet~\cite{russakovsky2015imagenet} for fair comparisons with other models.

\textbf{iBOT}~\cite{zhou2021ibot} can be viewed as the combination of self-distillation (DINO) and masked image modeling. Self-distillation is used to train a teacher network/online tokenizer that captures high-level visual semantics. Masked image modeling is used to train a student network that can recover each masked patch token to its corresponding teacher network/online tokenizer output.

\textbf{CLIP}~\cite{radford2021learning} is a simple and scalable method to learn visual representations with language supervision. CLIP forces the image representations to be aligned with paired captions through contrastive learning. The learned representations have strong zero-shot transferability and are effective for some robotic manipulation~\cite{shridhar2022cliport} and Embodied AI tasks~\cite{khandelwal2022simple}.

\section{Environments}
\label{app:envs}

We use 21 tasks across 3 robot manipulation environments: Meta-World (8 tasks), Robosuite (8 tasks), and Franka-Kitchen (5 tasks). We compare these 3 environments to other commonly used environments in Table~\ref{tab:env-comparison}.

\begin{table}[h!]
    \caption{Comparisons of different control environments. \textit{Kitchen} is short for Franka-Kitchen. Habitat does not support full-physics simulation and is only suitable for studying high-level planning. DeepMind Control (DMC) Suite focuses on locomotion tasks and the observations are not visually realistic. RLbench~\cite{james2020rlbench} has the most tasks but sparse rewards and low simulation speed.}
    \label{tab:env-comparison}
    \centering
    \small
    \begin{tabular}{lcccccc}
    \toprule
                  & Meta-World & Robosuite & Kitchen & Habitat & DMC & RLbench \\
    \midrule
    Full-physics &$\checkmark$&$\checkmark$&$\checkmark$& \ding{55} &$\checkmark$ &$\checkmark$ \\
    Visually-realistic &$\checkmark$&$\checkmark$&$\checkmark$ &$\checkmark$ & \ding{55} &$\checkmark$ \\
    Dense rewards &$\checkmark$&$\checkmark$&$\checkmark$ &$\checkmark$ &$\checkmark$ & \ding{55} \\
    Simulator     & MuJoCo & MuJoCo & MuJoCo & Habitat-Sim & MuJoCo & Coppelia\\
    \bottomrule
    \end{tabular}
\end{table}

\begin{figure*}[!t]
\captionsetup[subfigure]{labelformat=empty}
\captionsetup[subfigure]{aboveskip=2pt,belowskip=3pt}
\centering
\begin{subfigure}{0.2\textwidth}
  \centering
  \includegraphics[width=0.9\linewidth]{figures/metaworld/hammer-v2.png}
  \caption{Hammer}
\end{subfigure}
\begin{subfigure}{0.2\textwidth}
  \centering
  \includegraphics[width=0.9\linewidth]{figures/metaworld/drawer-close-v2.png}
  \caption{Drawer Close}
\end{subfigure}
\begin{subfigure}{0.2\textwidth}
  \centering
  \includegraphics[width=0.9\linewidth]{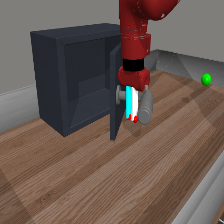}
  \caption{Door Open}
\end{subfigure}
\begin{subfigure}{0.2\textwidth}
  \centering
  \includegraphics[width=0.9\linewidth]{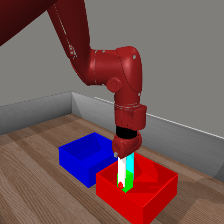}
  \caption{Bin Picking}
\end{subfigure} \\
\begin{subfigure}{0.2\textwidth}
  \centering
  \includegraphics[width=0.9\linewidth]{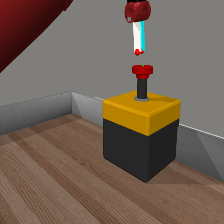}
  \caption{Button Press Topdown}
\end{subfigure}
\begin{subfigure}{0.2\textwidth}
  \centering
  \includegraphics[width=0.9\linewidth]{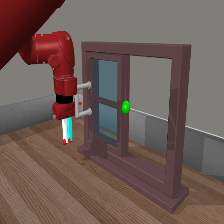}
  \caption{Window Close}
\end{subfigure}
\begin{subfigure}{0.2\textwidth}
  \centering
  \includegraphics[width=0.9\linewidth]{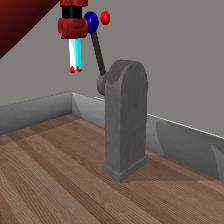}
  \caption{Lever Pull}
\end{subfigure}
\begin{subfigure}{0.2\textwidth}
  \centering
  \includegraphics[width=0.9\linewidth]{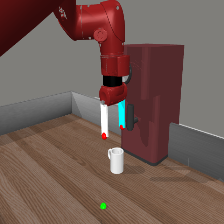}
  \caption{Coffee Pull}
\end{subfigure}
\vskip -0.1in
\caption{8 Meta-World tasks we consider in our study.}
\label{fig:metaworld_env}
\end{figure*}

\textbf{Meta-World}~\cite{yu2020meta} is a simulated environment for multi-task learning and meta-learning consisting of 50 robotic manipulation tasks. We select 8 distinct tasks: \texttt{hammer}, \texttt{drawer-close}, \texttt{door-open}, \texttt{bin-picking}, \texttt{button-press-topdown}, \texttt{window-close}, \texttt{lever-pull} and \texttt{coffee-pull}, depicted in Figure~\ref{fig:metaworld_env}. All the tasks are performed by a simulated Sawyer robot. The episode length for each task is 125 steps, except for \texttt{bin-picking} and \texttt{lever-pull}, which run for 175 steps. The positions of the target objects (e.g., hammer, drawer, door, etc.) are randomized between episodes. For each task, we use the task-specific hard-coded policies provided in the open-source implementation to collect a total of 25 expert demonstrations. Our BC experiments consider a varying number of expert demonstrations: [1, 5, 10, 15, 20, 25], but for all the VRF experiments, we only use 1 demonstration.

\begin{figure*}[!t]
\captionsetup[subfigure]{labelformat=empty}
\captionsetup[subfigure]{aboveskip=2pt,belowskip=3pt}
\centering
\begin{subfigure}{0.2\textwidth}
  \centering
  \includegraphics[width=0.9\linewidth]{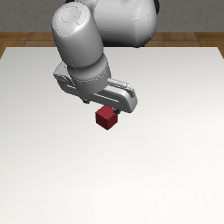}
  \caption{Panda Lift}
\end{subfigure}
\begin{subfigure}{0.2\textwidth}
  \centering
  \includegraphics[width=0.9\linewidth]{figures/robosuite/Panda_Door.png}
  \caption{Panda Door}
\end{subfigure}
\begin{subfigure}{0.2\textwidth}
  \centering
  \includegraphics[width=0.9\linewidth]{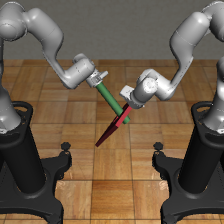}
  \caption{Panda Two Arm}
\end{subfigure}
\begin{subfigure}{0.2\textwidth}
  \centering
  \includegraphics[width=0.9\linewidth]{figures/robosuite/PickPlaceCan.png}
  \caption{Panda Pick-and-Place}
\end{subfigure}
\begin{subfigure}{0.2\textwidth}
  \centering
  \includegraphics[width=0.9\linewidth]{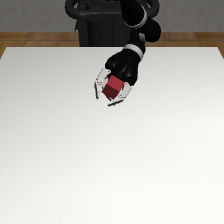}
  \caption{Jaco Lift}
\end{subfigure}
\begin{subfigure}{0.2\textwidth}
  \centering
  \includegraphics[width=0.9\linewidth]{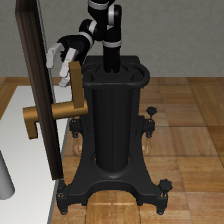}
  \caption{Jaco Door}
\end{subfigure}
\begin{subfigure}{0.2\textwidth}
  \centering
  \includegraphics[width=0.9\linewidth]{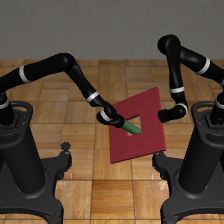}
  \caption{Jaco Two Arm}
\end{subfigure}
\begin{subfigure}{0.2\textwidth}
  \centering
  \includegraphics[width=0.9\linewidth]{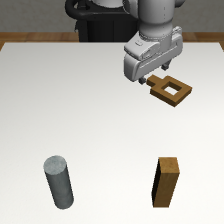}
  \caption{Panda Nut Assembly}
\end{subfigure}
\vskip -0.1in
\caption{8 Robosuite tasks we consider in our study.}
\label{fig:robosuite_env}
\end{figure*}

\textbf{Robosuite}~\cite{zhu2020robosuite} is a modular simulation framework and benchmark for robot learning. We consider 8 tasks from the environment, including \texttt{Panda-Lift}, \texttt{Panda-Door}, \texttt{Panda-TwoArmPegInHole}, \texttt{Panda-PickPlaceCan}, \texttt{Panda-NutAssemblySquare}, \texttt{Jaco-Lift}, \texttt{Jaco-Door} and \texttt{Jaco-TwoArmPegInHole}, as illustrated in Figure~\ref{fig:robosuite_env}. 5 out of 8 tasks are performed by a Panda robot with a parallel-jaw gripper, while the other 3 tasks are performed by a Jaco robot with multi-jointed fingers, which are harder to control. We use Operational Space Controllers (OSC)~\cite{khatib1995inertial} to transform the high-level actions into low-level virtual motor commands. The episode length for all tasks is 80 steps (except \texttt{Panda-PickPlaceCan} and \texttt{Panda-NutAssemblySquare}, which runs for 150 and 160 steps, respectively). In all tasks, the locations of objects (e.g., cube, can, nut, etc.) are randomized at the beginning of each episode. For \texttt{Panda-PickPlaceCan} and \texttt{Panda-NutAssemblySquare}, we use 50 expert demonstrations provided by robomimic~\cite{robomimic2021}. For the other tasks, we train a state-based DrQ-v2 and collect 50 demonstrations for each task. Our BC experiments consider a different number of expert demonstrations:  [5, 10, 20, 30, 40, 50], but all the VRF experiments only use 1 demonstration.

\begin{figure*}[!t]
\captionsetup[subfigure]{labelformat=empty}
\captionsetup[subfigure]{aboveskip=2pt,belowskip=3pt}
\centering
\begin{subfigure}{0.19\textwidth}
  \centering
  \includegraphics[width=0.9\linewidth]{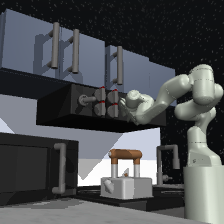}
  \caption{Turning Knob}
\end{subfigure}
\begin{subfigure}{0.19\textwidth}
  \centering
  \includegraphics[width=0.9\linewidth]{figures/kitchen/kitchen_micro_open-v3.png}
  \caption{Opening Microwave}
\end{subfigure}
\begin{subfigure}{0.19\textwidth}
  \centering
  \includegraphics[width=0.9\linewidth]{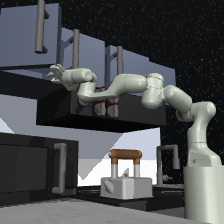}
  \caption{Opening Door}
\end{subfigure}
\begin{subfigure}{0.19\textwidth}
  \centering
  \includegraphics[width=0.9\linewidth]{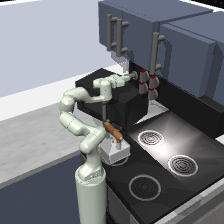}
  \caption{Turning Light On}
\end{subfigure}
\begin{subfigure}{0.19\textwidth}
  \centering
  \includegraphics[width=0.9\linewidth]{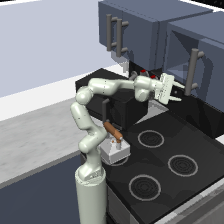}
  \caption{Sliding Door}
\end{subfigure}

\vskip -0.1in
\caption{5 Franka-Kitchen tasks we consider in our study.}
\label{fig:kitchen_env}
\end{figure*}

\textbf{Franka-Kitchen}~\cite{gupta2019relay} requires to control a 9 DoF Franka robot to perform various tasks in a kitchen scene. In this study, we consider 5 tasks: \texttt{knob1-on}, \texttt{micro-open}, \texttt{ldoor-open}, \texttt{light-on} and \texttt{sdoor-open}, as depicted in Figure~\ref{fig:kitchen_env}. The episode length for all Franka tasks is 50 steps. Following PVR~\cite{parisi2022unsurprising}, we randomize the pose of the robot arm between episodes but not the scene itself. We train expert policies using state-based DrQ-v2 and collect 25 expert demonstrations for each task. Our BC experiments consider a varying number of expert demonstrations: [1, 5, 10, 15, 20, 25], but for all VRF experiments, we only use 1 demonstration.

\section{Implementation Details}
\label{app:details}

\subsection{Policy Learning}
\label{app:policy-details}

For all the environments and tasks, the observations are 224 × 224 RGB images with no access to proprioceptive information. In our experiments, we find that using only one image observation is comparable to using a stack of consecutive images, so we choose the more compute-efficient option. Additionally, we do not use image augmentation (e.g., random shift). For all Meta-World and Robosuite tasks, we employ action repeat of 2, while for Franka-Kitchen tasks, the action repeat is set to 1 (no action repeat).
Further implementation details for each policy learning method are as follows.

\textbf{Reinforcement learning.}
All the policies are trained for 3M environment steps. We set the replay buffer size to 500000 and increase the mini-batch size to 512. 
The features output by a pre-trained vision model are fed into the actor and critic networks, whose architectures follow DrQ-v2: a `trunk' network and a small MLP network. 
The `trunk' network is a single fully-connected layer with \texttt{LayerNorm}~\cite{ba2016layer} and \texttt{tanh} nonlinearity. The output dimension of the `trunk' network is 50, forming a bottleneck structure. 
The MLP network has 3 layers, and the hidden dimension is set to 1024.
The complete list of hyper-parameters is in Table~\ref{tab:rl-hyper}. 

\textbf{Imitation learning through behavior cloning.}
For BC, we discard the bottleneck structure in DrQ-v2's actor and use an architecture as follows: a four-layer MLP with hidden sizes [512, 256, 128] and \texttt{ReLU} activations. The MLP network is preceded by a \texttt{LayerNorm} layer to calibrate the feature magnitudes across different pre-trained vision models. We train the policy with mini-batches of 128 samples for 110000 steps with the Adam optimizer~\cite{kingma2014adam} (learning rate 0.0001).

\textbf{Imitation learning with a visual reward function.}
We closely follow the implementation of ROT. All the policies are trained for 1M environment steps. As we can compute the imitation rewards on the output features of pre-trained vision models, we no longer need the target feature processor in the original ROT. See Table~\ref{tab:vrf-hyper} for all the hyper-parameters.

\begin{table}[h!]
\caption{A default set of hyper-parameters for RL.}
\label{tab:rl-hyper} 
\centering
\begin{tabular}{lc}
\toprule
Config  & Value \\
\midrule
Training environment steps & $3.1 \times 10^6$ \\
Replay buffer capacity & $500000$ \\
Seed frames & $4000$ \\
Exploration steps & $2000$ \\
$n$-step returns & $3$ \\
Mini-batch size & $512$ \\
Discount $\gamma$ & $0.99$ \\
Optimizer & Adam \\
Learning rate & $10^{-4}$ \\
Agent update frequency & $2$ \\
Critic Q-function soft-update rate $\tau$ & $0.01$ \\
Features dim. & $50$ \\
Hidden dim. & $1024$ \\
Exploration stddev. clip & $0.3$ \\
Exploration stddev. schedule & $\mathrm{linear}(1.0, 0.1, 800000)$ \\
\bottomrule
\end{tabular}
\end{table}

\begin{table}[h!]
\caption{A default set of hyper-parameters for VRF.}
\label{tab:vrf-hyper} 
\centering
\begin{tabular}{lc}
\toprule
Config  & Value \\
\midrule
Training environment steps & $1.1 \times 10^6$ \\
Replay buffer capacity & $150000$ \\
Seed frames & $12000$ \\
Exploration steps & $0$ \\
$n$-step returns & $3$ \\
Mini-batch size & $256$ \\
Discount $\gamma$ & $0.99$ \\
Optimizer & Adam \\
Learning rate & $10^{-4}$ \\
Agent update frequency & $2$ \\
Critic Q-function soft-update rate $\tau$ & $0.01$ \\
Features dim. & $50$ \\
Hidden dim. & $1024$ \\
Exploration stddev. clip & $0.3$ \\
Exploration stddev. schedule & 0.1 \\
BC weight type & qfilter \\
Auto reward scale factor & $10$ \\
\bottomrule
\end{tabular}
\end{table}

\subsection{Linear Probing}
\label{app:probing-details}

We train a linear regressor on Franka-Kitchen image features output by frozen pre-trained vision models to predict the ground-truth environment features, which are composed of joint positions, object locations, end-effector poses, and relative distances between object and end-effector. The dimension of this environment feature is 57. We adopt an extra \texttt{LayerNorm} layer before the linear regressor to calibrate the feature magnitudes across different vision models. We use $lr=0.02$ with a warmup of 10 epochs and cosine learning rate decay, weight decay $=0$, and batch size $=16$ with an Adam optimizer. We train on 40 expert demonstrations (2000 images) for 100 epochs and report validation loss on 10 expert demonstrations.

\section{Additional Experimental Results}

\subsection{Reinforcement Learning}
\label{app:rl-uncertainty}

\begin{figure*}[h!]
    \centering
    \includegraphics[width=0.8\linewidth]{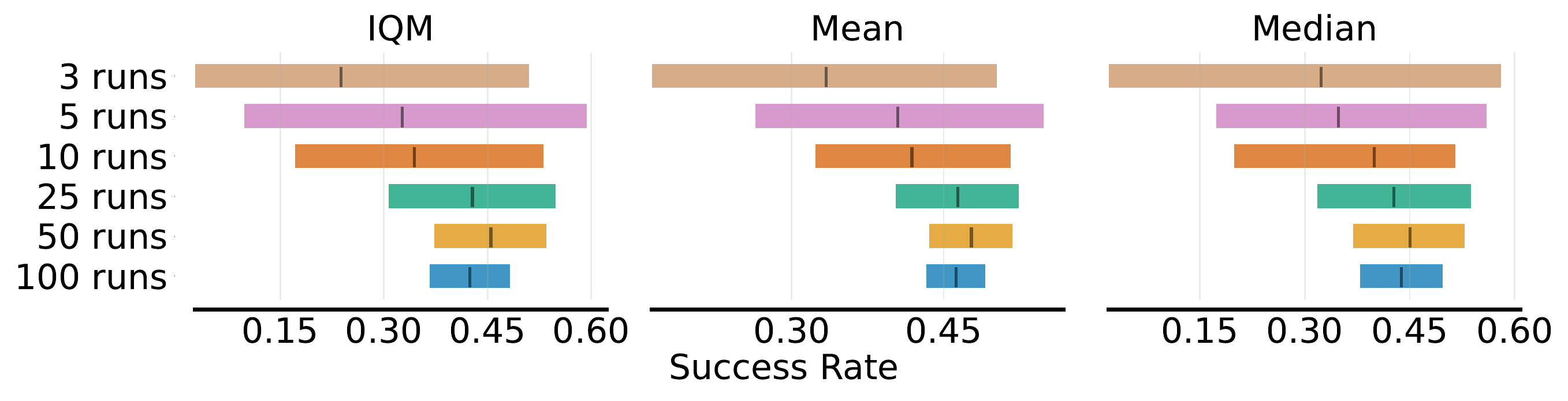} 
    \caption{RL performance with 95\% CIs for \textit{a varying number of runs} for IQM, Mean and Median scores for R3M. We use 5 tasks from Meta-World: \texttt{hammer}, \texttt{door-open}, \texttt{bin-picking}, \texttt{button-press-topdown}, \texttt{lever-pull} and train all the RL agents for 2M environment steps. The other experimental settings are the same as described in Appendix~\ref{app:policy-details}.}
    \label{fig:r3m-100seeds}
\end{figure*}

Folk wisdom in experimental RL suggests that evaluating more runs per task can reduce uncertainty. In this section, we investigate the following question: ``How many runs with different seeds are required to obtain stable and reliable results?'' By observing the 95\% confidence intervals in Figure~\ref{fig:r3m-100seeds}, we find that there is substantial uncertainty in scores even with 50 runs and it may be necessary to perform 100 or more runs to address the issue of statistical uncertainty effectively.

\begin{figure*}[h!]
    \centering
    \includegraphics[width=0.5\linewidth]{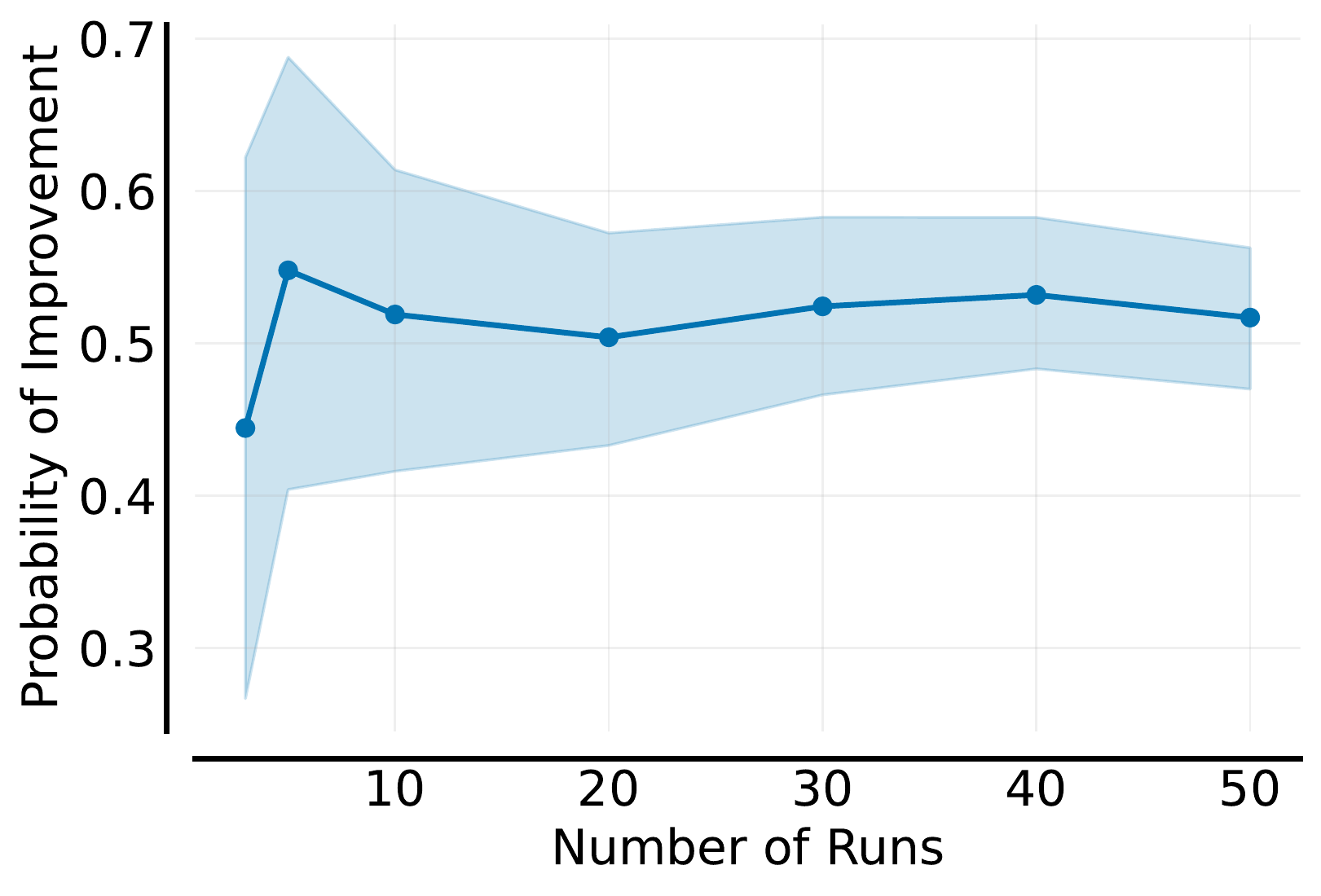} 
    \caption{The probability of improvement, with 95\% CIs (shaded regions), that R3M outperforms VFS, for \textit{a varying number of runs}. The interval estimates are based on stratified bootstrap with independent sampling with 2000 bootstrap re-samples. We consider 5 tasks (the same as Fig.~\ref{fig:r3m-100seeds}) from Meta-World and train all the RL agents for 2M environment steps. The other settings are the same as described in Appendix~\ref{app:policy-details}.}
    \label{fig:50seeds-probability}
\end{figure*}

Next, we ask the question: ``How many training runs do we need to compare two pre-trained vision models reliably?'' We show the \textit{probability of improvement}~\cite{agarwal2021deep} of R3M compared to VFS for a varying number of runs in Figure~\ref{fig:50seeds-probability}. With a handful of runs (i.e., 3 or 5), the confidence intervals are substantially large. The comparison is stable and reliable only when 40 or more runs are used, however, this number of runs is often computationally prohibitive for complex and challenging benchmarks.

\subsection{Full Aggregate Metrics on BC and VRF}
\label{app:additioal-exp}

\begin{figure*}[!h]
\captionsetup[subfigure]{labelformat=empty}
\centering
\begin{subfigure}{0.85\textwidth}
  \centering
  \caption{ResNet-50}
  \includegraphics[width=\linewidth]{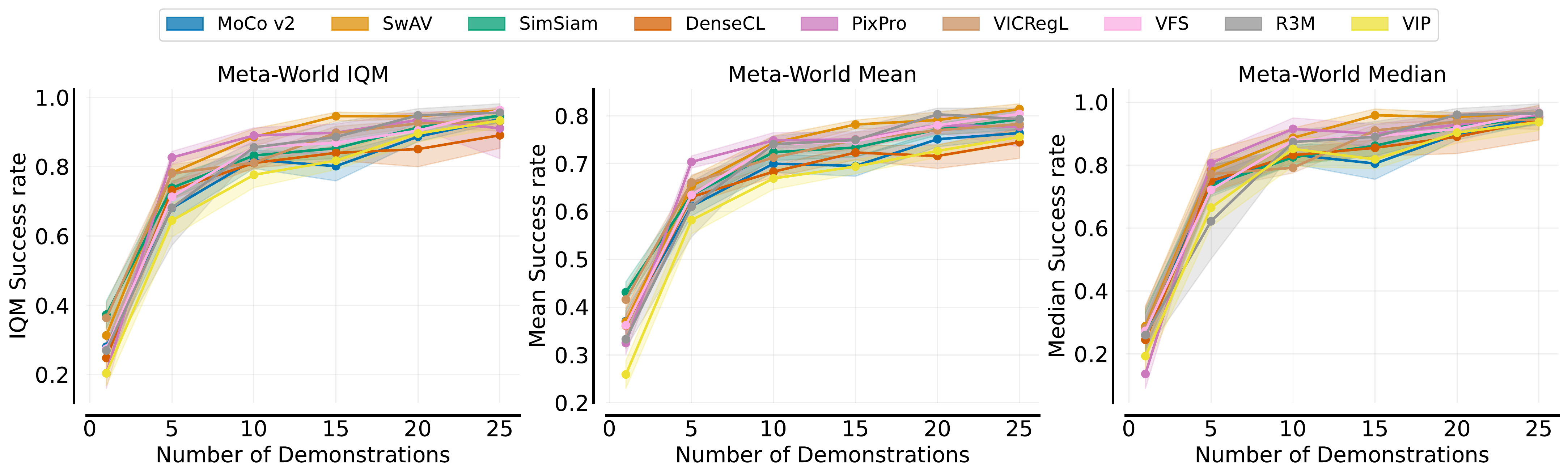}
\end{subfigure} \\
\begin{subfigure}{0.85\textwidth}
  \centering
  \caption{ViT-B/16}
  \includegraphics[width=\linewidth]{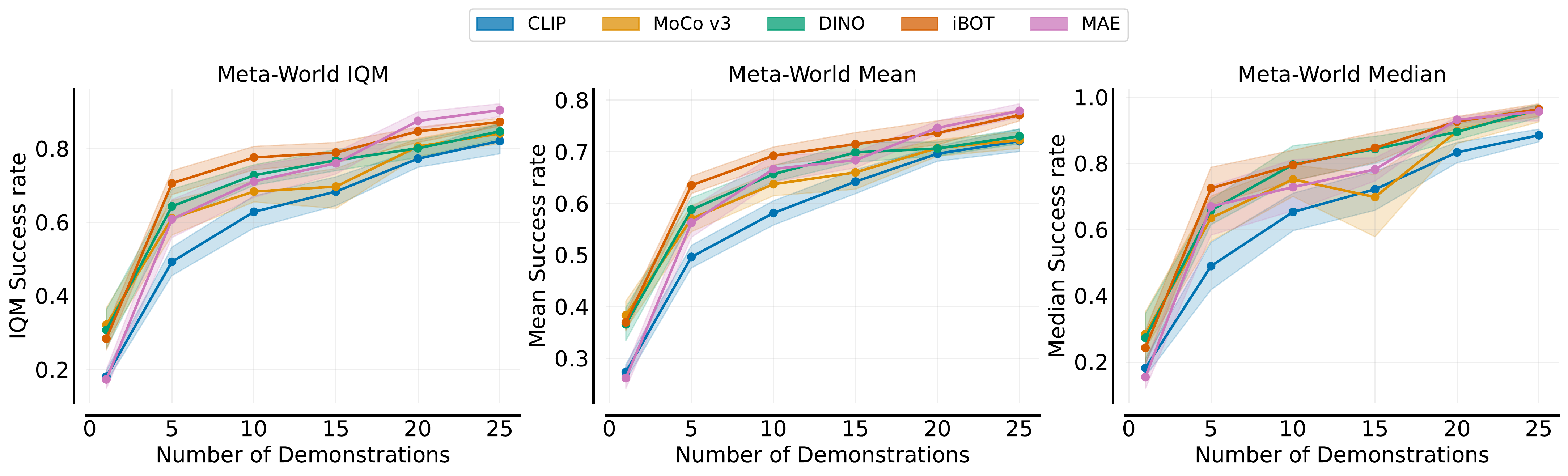}
\end{subfigure}
\caption{BC performance across 3 aggregate metrics (IQM, Mean, and Median) on Meta-World. Shaded regions show pointwise 95\% percentile stratified bootstrap CIs.}
\label{fig:bc-results-metaworld}
\end{figure*}

\begin{figure*}[!t]
\captionsetup[subfigure]{labelformat=empty}
\centering
\begin{subfigure}{0.85\textwidth}
  \centering
  \caption{ResNet-50}
  \includegraphics[width=\linewidth]{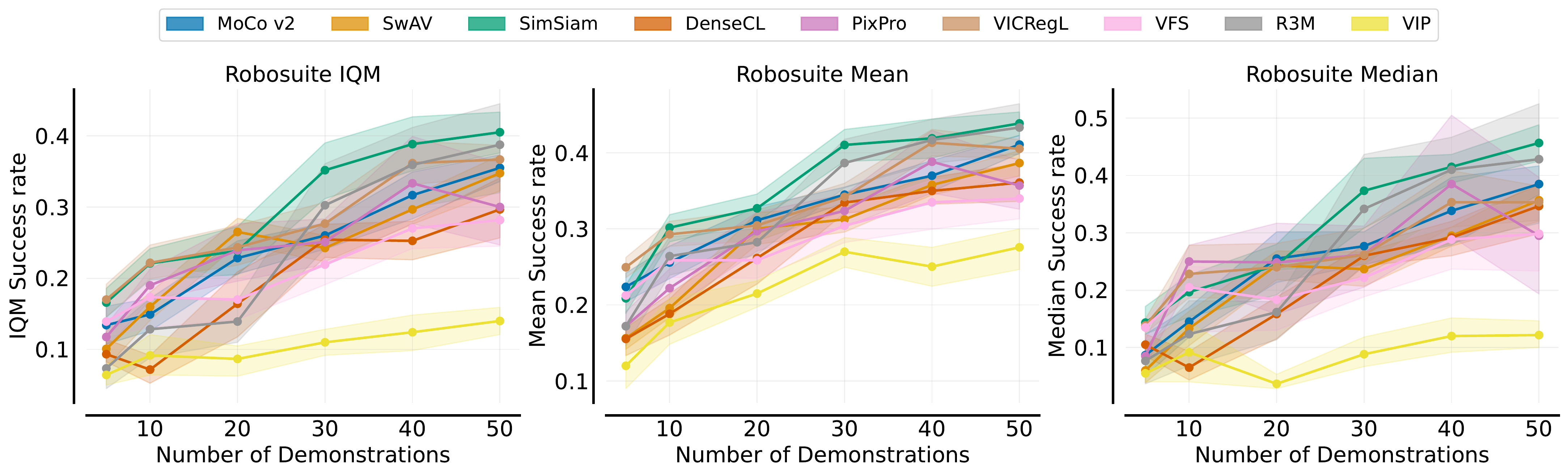}
\end{subfigure} \\
\begin{subfigure}{0.85\textwidth}
  \centering
  \caption{ViT-B/16}
  \includegraphics[width=\linewidth]{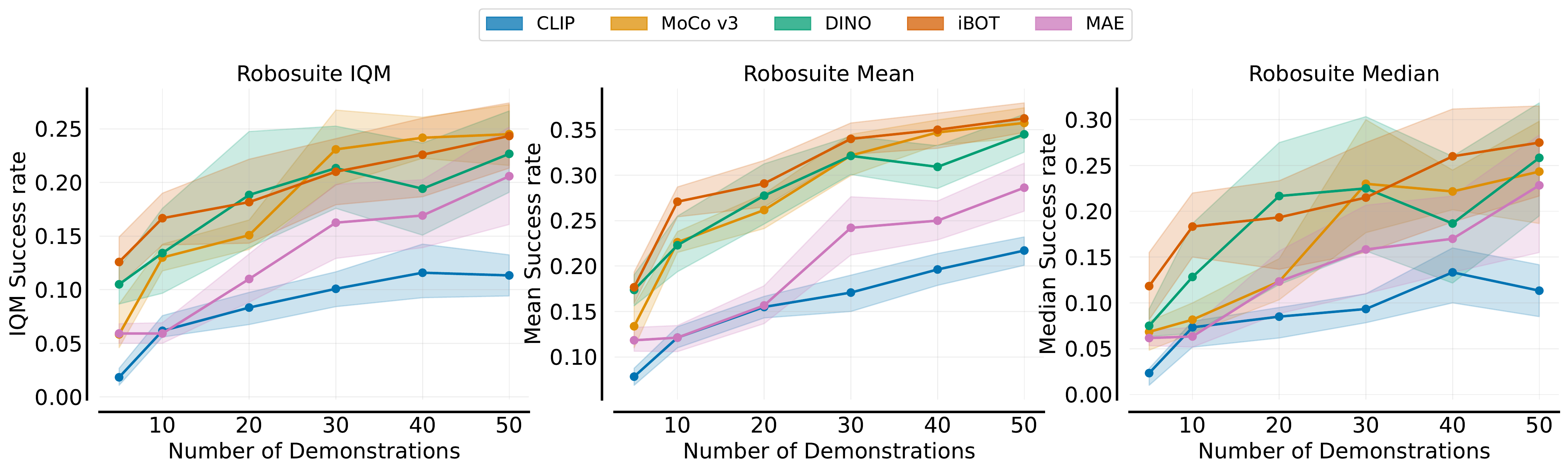}
\end{subfigure}
\caption{BC performance across 3 aggregate metrics (IQM, Mean, and Median) on Robosuite. Shaded regions show pointwise 95\% percentile stratified bootstrap CIs.}
\label{fig:bc-results-robosuite}
\end{figure*}

\begin{figure*}[!t]
\captionsetup[subfigure]{labelformat=empty}
\centering
\begin{subfigure}{0.85\textwidth}
  \centering
  \caption{ResNet-50}
  \includegraphics[width=\linewidth]{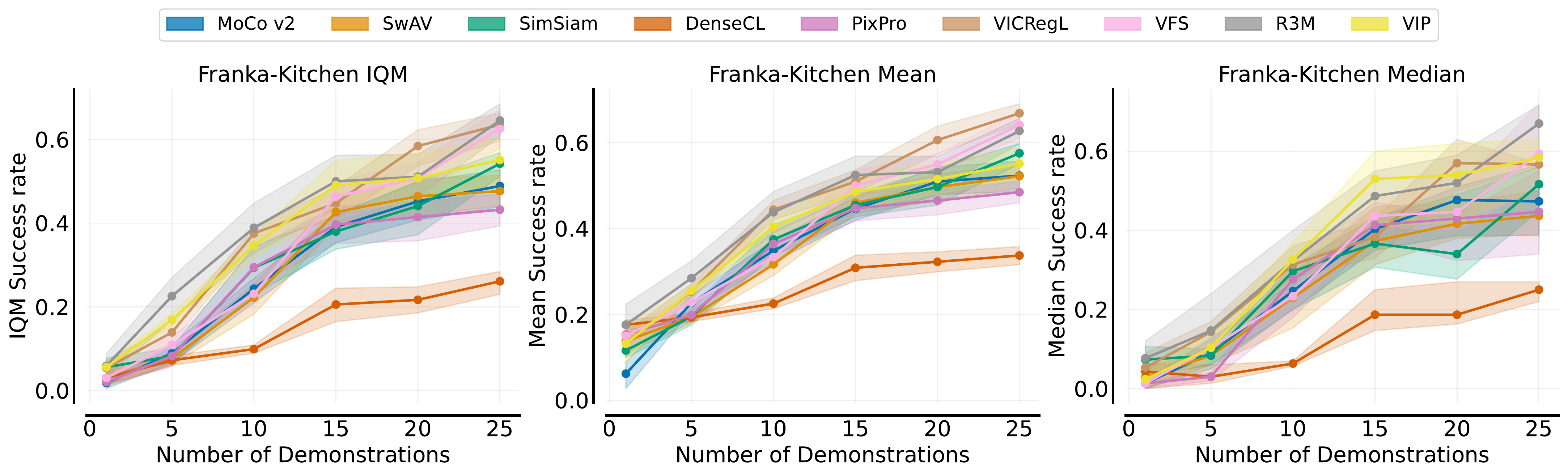}
\end{subfigure} \\
\begin{subfigure}{0.85\textwidth}
  \centering
  \caption{ViT-B/16}
  \includegraphics[width=\linewidth]{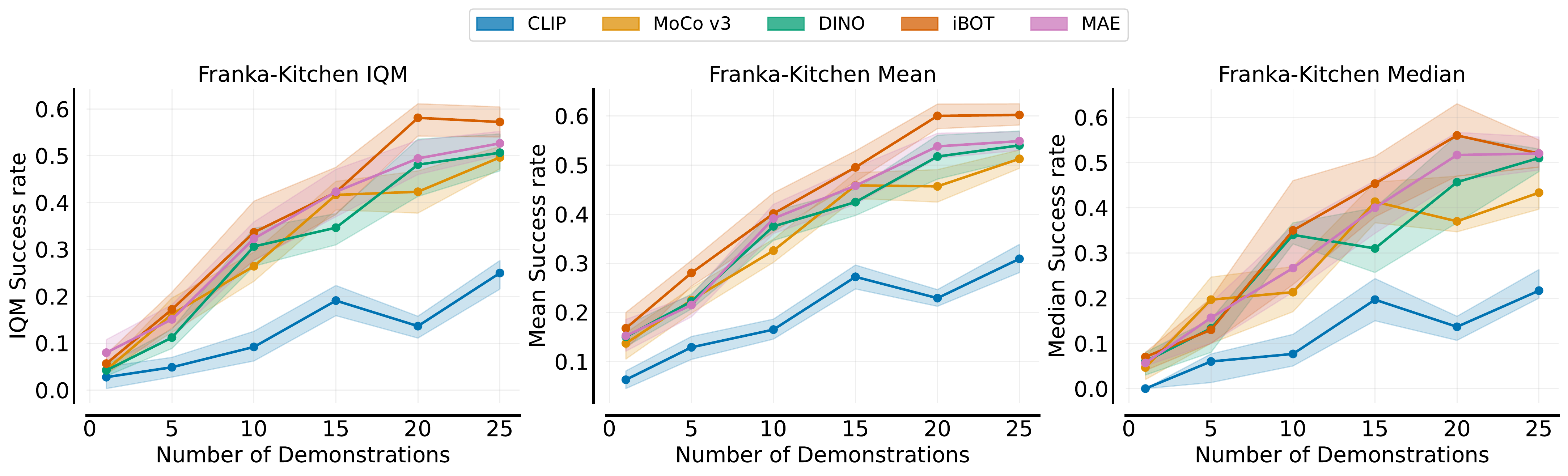}
\end{subfigure}
\caption{BC performance across 3 aggregate metrics (IQM, Mean, and Median) on Franka-Kitchen. Shaded regions show pointwise 95\% percentile stratified bootstrap CIs.}
\label{fig:bc-results-kitchen}
\end{figure*}

\begin{figure*}[!t]
\captionsetup[subfigure]{labelformat=empty}
\centering
\begin{subfigure}{0.85\textwidth}
  \centering
  \caption{ResNet-50}
  \includegraphics[width=\linewidth]{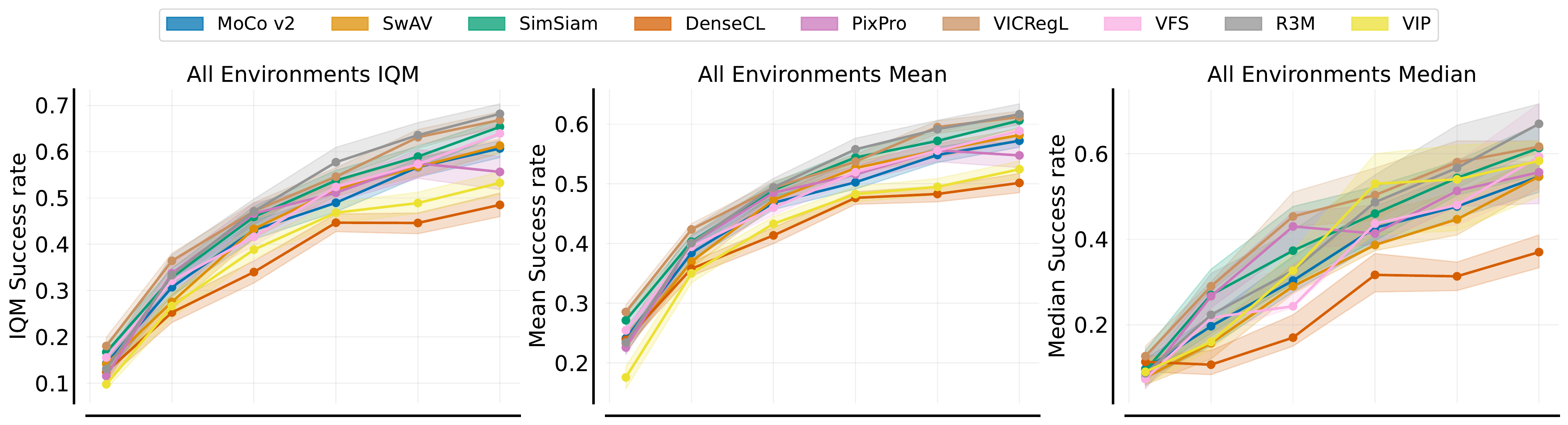}
\end{subfigure} \\
\begin{subfigure}{0.85\textwidth}
  \centering
  \caption{ViT-B/16}
  \includegraphics[width=\linewidth]{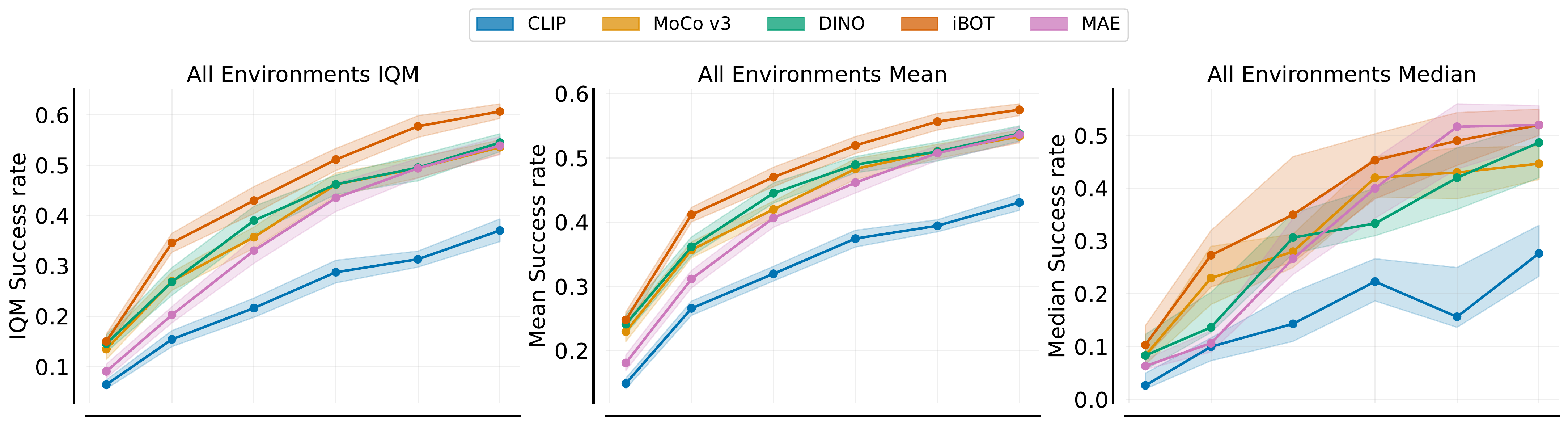}
\end{subfigure}
\caption{BC performance across 3 aggregate metrics (IQM, Mean, and Median) on 21 tasks of all environments. Shaded regions show pointwise 95\% percentile stratified bootstrap CIs.}
\label{fig:bc-results-allenv}
\end{figure*}

Figure~\ref{fig:bc-results-metaworld}, Figure~\ref{fig:bc-results-robosuite}, Figure~\ref{fig:bc-results-kitchen} and Figure~\ref{fig:bc-results-allenv} show the BC results across 3 aggregate metrics (IQM, Mean, Median) on Meta-World, Robosuite, Franka-Kitchen and all environements, respectively.

\begin{figure*}[t!]
    \centering
    \includegraphics[width=0.8\linewidth]{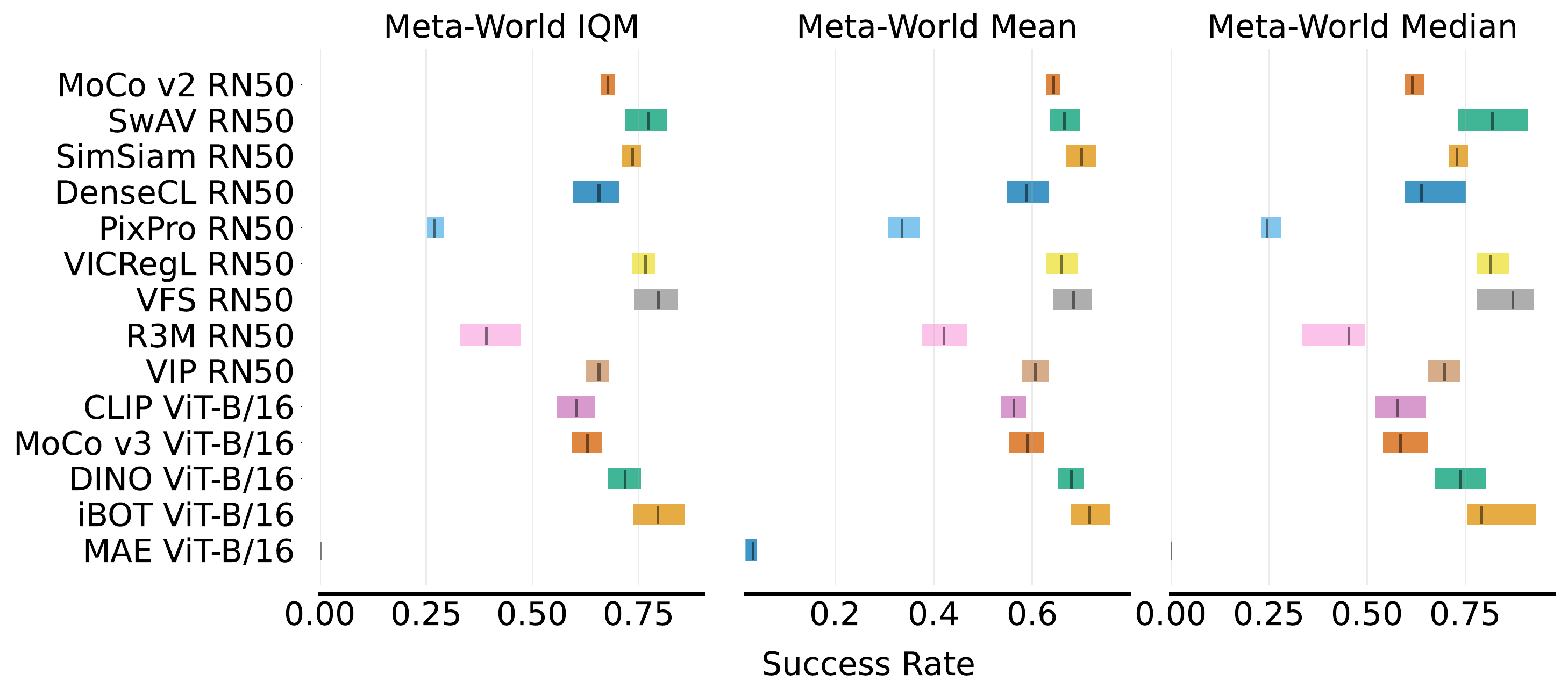} 
    \caption{VRF performance across 3 aggregate metrics (IQM, Mean, and Median) on Meta-World with 95\% CIs.}
    \label{fig:rot-results-metaworld}
\end{figure*}

\begin{figure*}[t!]
    \centering
    \includegraphics[width=0.8\linewidth]{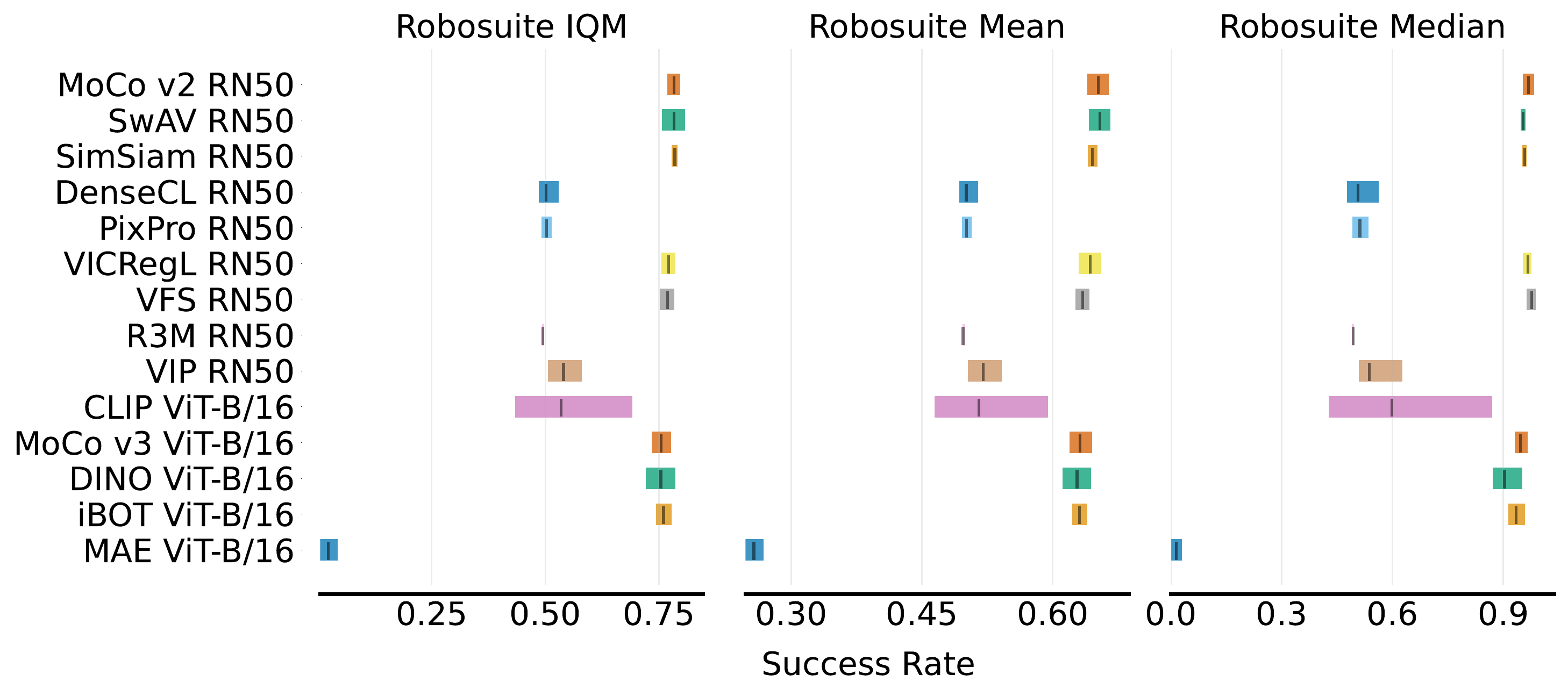} 
    \caption{VRF performance across 3 aggregate metrics (IQM, Mean, and Median) on Robosuite with 95\% CIs..}
    \label{fig:rot-results-robosuite}
\end{figure*}

\begin{figure*}[t!]
    \centering
    \includegraphics[width=0.8\linewidth]{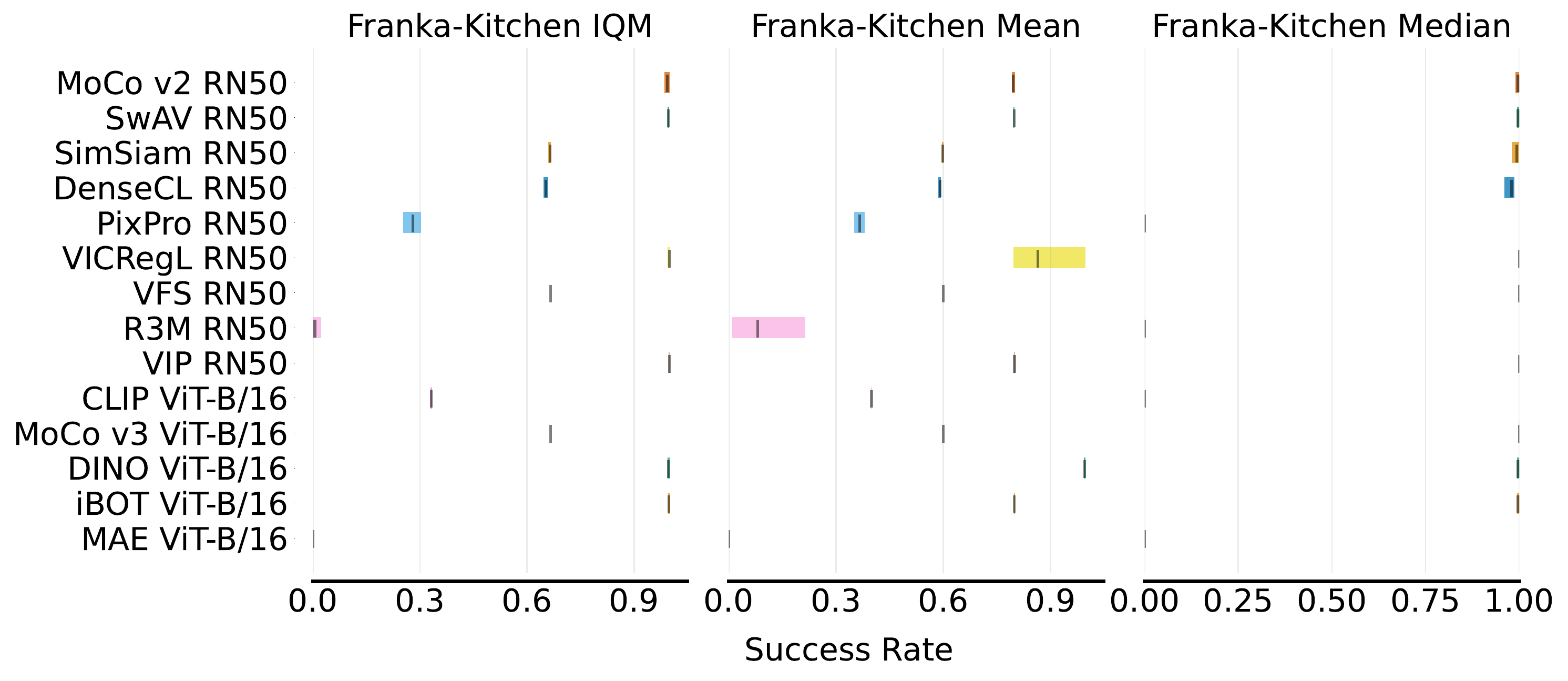} 
    \caption{VRF performance across 3 aggregate metrics (IQM, Mean, and Median) on Franka-Kitchen with 95\% CIs.}
    \label{fig:rot-results-kitchen}
\end{figure*}

\begin{figure*}[t!]
    \centering
    \includegraphics[width=0.8\linewidth]{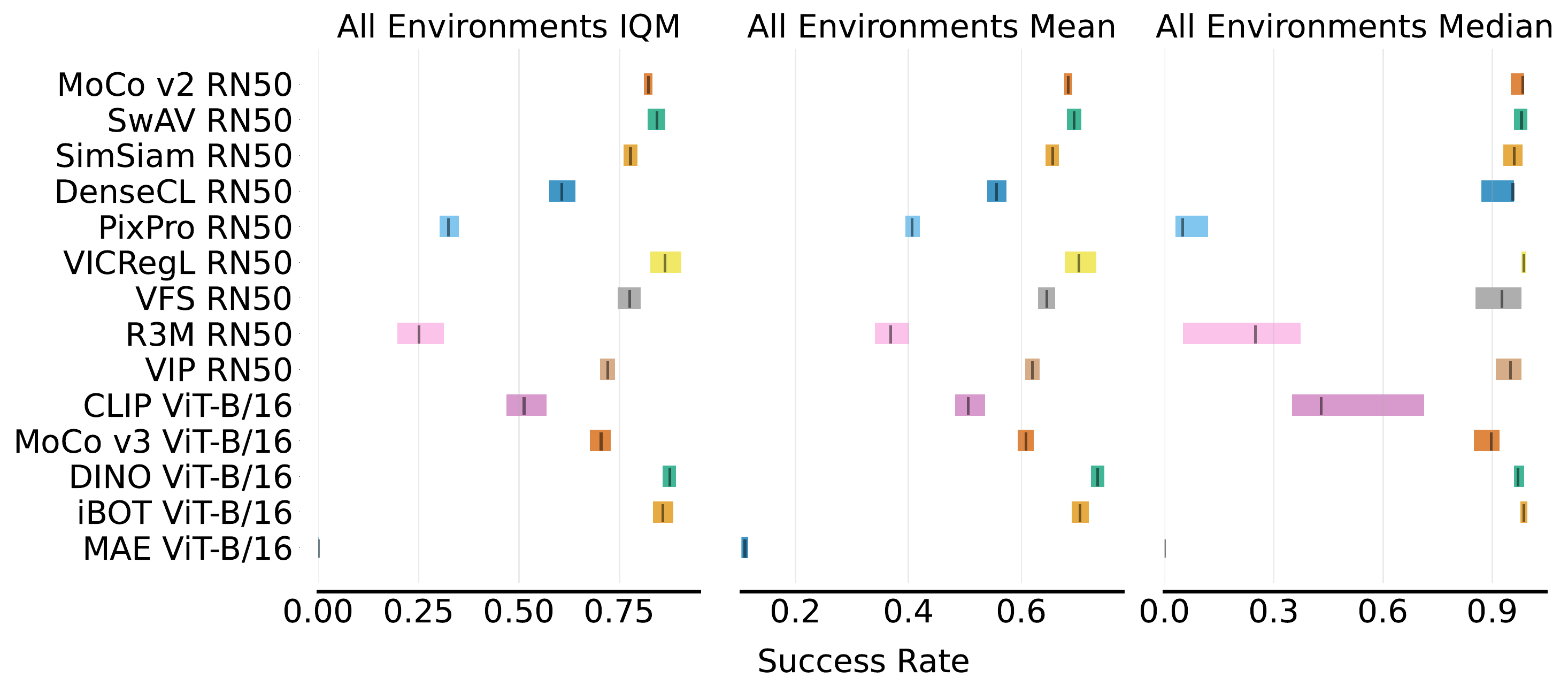} 
    \caption{VRF performance across 3 aggregate metrics (IQM, Mean, and Median) on 21 tasks of all environments with 95\% CIs.}
    \label{fig:rot-results-allenv}
\end{figure*}

Figure~\ref{fig:rot-results-metaworld}, Figure~\ref{fig:rot-results-robosuite}, Figure~\ref{fig:rot-results-kitchen} and Figure~\ref{fig:rot-results-allenv} show the VRF results across 3 aggregate metrics (IQM, Mean, Median) on Meta-World, Robosuite, Franka-Kitchen and all environements, respectively.

\end{document}